\documentclass{article}

\usepackage[accepted]{aistats2021}

\usepackage{amssymb}

\usepackage[dvipsnames]{xcolor}
\usepackage[hyperfootnotes=false, hidelinks]{hyperref}
\hypersetup{ %
    pdftitle={},
    pdfauthor={},
    colorlinks=true,
    linkcolor=MidnightBlue,
    citecolor=MidnightBlue,
    filecolor=MidnightBlue,
    urlcolor=MidnightBlue,
    }

\usepackage{amsmath,amsthm}
\usepackage{graphicx}

\usepackage{subfig}
  
\usepackage{wrapfig}

\usepackage{multirow}

\numberwithin{equation}{section}

\usepackage[round]{natbib}

\usepackage{appendix}

\usepackage{booktabs} 

\DeclareMathOperator{\expect}{\mathbb{E}}

\DeclareMathOperator{\ELBO}{\mathcal{L}}
\DeclareMathOperator{\KL}{\mathrm{KL}}

\newcommand{\vecto}[1]{\boldsymbol{\mathbf{#1}}}
\renewcommand{\v}{\vecto}

\newcommand{\beq}{\begin{eqnarray}}
\newcommand{\eeq}{\end{eqnarray}}
\newcommand{\beqs}{\begin{eqnarray*}}
\newcommand{\eeqs}{\end{eqnarray*}}

\DeclareMathOperator*{\argmax}{arg\max}

\newtheorem{theorem}{Theorem}
 
\newtheorem{proposition}[theorem]{Proposition}

\begin{document}

\twocolumn[
\aistatstitle{Variational Autoencoders: A Harmonic Perspective}

\aistatsauthor{Alexander Camuto \And Matthew Willetts}

\aistatsaddress{ University of Oxford \And  UCL  \& The Alan Turing Institute} ]



\begin{abstract}
In this work we study Variational Autoencoders (VAEs) from the perspective of harmonic analysis. 
By viewing a VAE's latent space as a Gaussian Space, a variety of measure space, we derive a series of results that show that the encoder variance of a VAE controls the frequency content of the functions parameterised by the VAE encoder and decoder neural networks. 
In particular we demonstrate that larger encoder variances reduce the high frequency content of these functions. 
Our analysis allows us to show that increasing this variance effectively induces a soft Lipschitz constraint on the decoder network of a VAE, which is a core contributor to the adversarial robustness of VAEs. 
We further demonstrate that adding Gaussian noise to the input of a VAE allows us to more finely control the frequency content and the Lipschitz constant of the VAE encoder networks. 
Finally, we show that the KL term of the VAE loss serves as single point of action for modulating the frequency content of both encoder and decoder networks; 
whereby upweighting this term decreases the high-frequency content of both networks. 
To support our theoretical analysis we run experiments using VAEs with small fully-connected neural networks and with larger convolutional networks, demonstrating empirically that our theory holds for a variety of neural network architectures. 
\end{abstract}

\section{Introduction}
\label{intro}

Variational autoencoders (VAEs) are deep latent variable models that typically use Gaussian priors and Gaussian posteriors in their latent spaces ~\citep{Rezende2014dgm, Kingma2013}. 
VAEs have become a work-horse method in modern machine learning, but still their theoretical properties are not fully understood.
In particular, we do not yet fully understand the regularising effect of latent space sampling during training.

While the effect of the latent space sampling (a.k.a latent noise) on  VAEs has been studied from an information-theoretic standpoint~\citep{shu2019reg} and through Taylor analysis~\citep{Kumar2020}, here we take a different tack and study the impact that latent Gaussian samples have on the \textit{harmonic properties} of the underlying neural networks used to implement the model.

Related to our inquiry is the study of the \textit{Gaussian noise injections} in neural networks trained on supervised tasks~\citep{yin2019fourier, Camuto2020_2, camuto2021asymmetric}. 
Adding (standard) Gaussian noise to the input layer has long been known to induce regularisation~\citep{Webb1994, Bishop1995, Burger2003}, and recently has been used to induce robustness to adversarial attacks~\citep{Cohen2019}.
Recently, by studying the effects of these Gaussian noise injections from a functional analysis perspective, \citep{Camuto2020_2} shows that these injections penalise functions which learn high frequency components in Fourier-space. 
These ideas form the starting point for our work.

By considering the latent space of a VAE as a measure space, equipped with a \textit{Gaussian measure}, we can consider the decoder of the VAE as being a member of a \textit{Gaussian Space}, a type of $L_2$ function space equipped with the Gaussian measure. 
In our analysis, we view the latent variable's posterior, a Gaussian, as being the Gaussian measure with which the latent space is equipped on a per-datapoint basis.  
This posterior is located around a particular \textit{location}, the mean, with a particular \textit{scale}, the (often-diagonal) standard deviation. 
\begin{figure*}[t!]
\centering
    \includegraphics[width=0.32\textwidth]{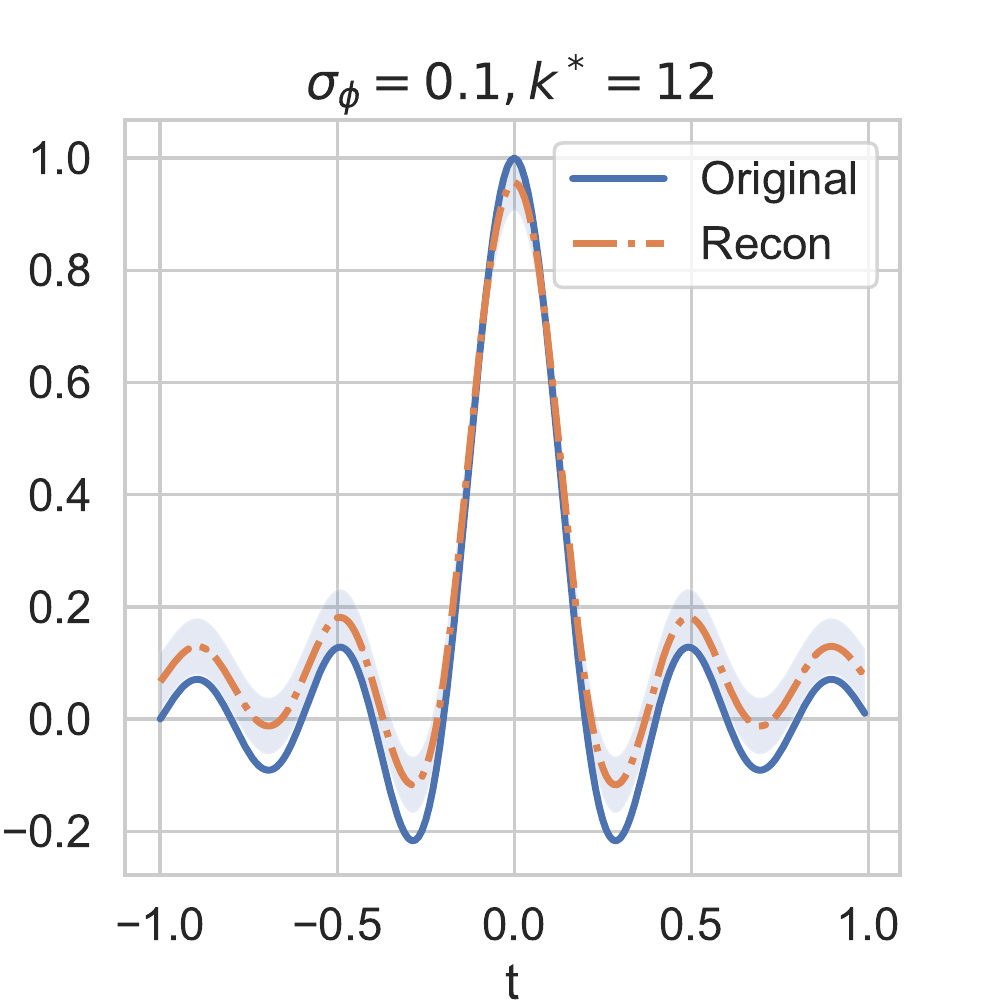}
    \includegraphics[width=0.32\textwidth]{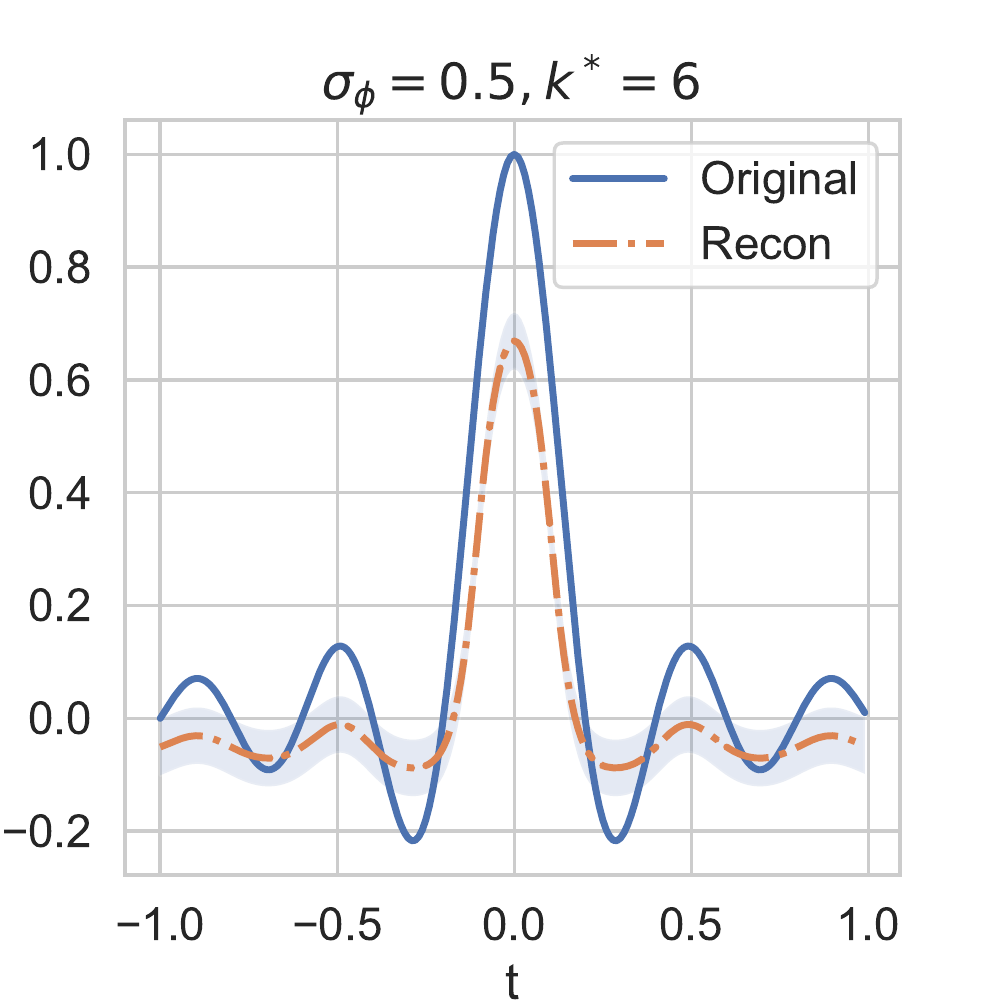}
    \includegraphics[width=0.32\textwidth]{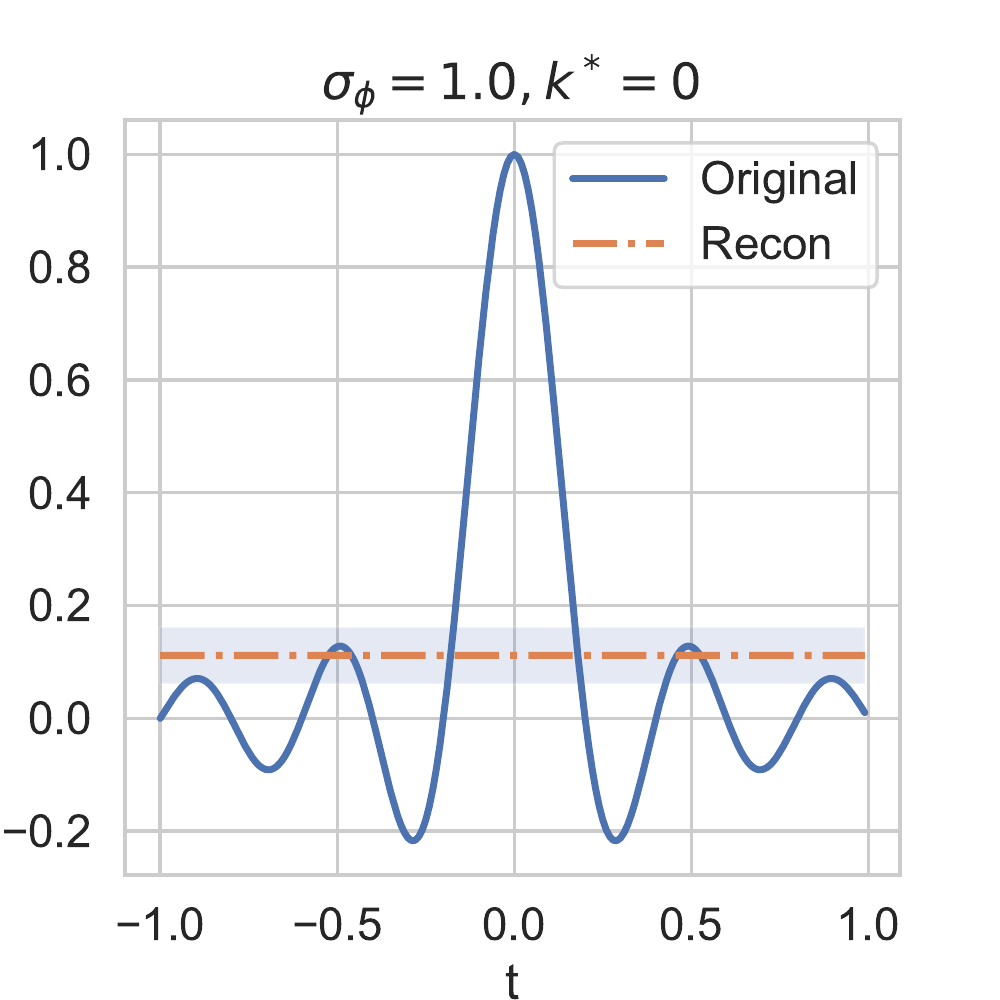}
    \caption{We plot the reconstructed function of a VAE, with the $\mathrm{Sigmoid}$ activation function networks (3 dense layer each with 256 units), trained on the function $\mathrm{sinc}(5t)$, $t \in [-1,1]$. We use a one dimensional latent space $\mathcal{Z}$ with a fixed encoder variance $\sigma_\phi \in [0.1,0.5,1.0]$. 
    As $\sigma_\phi$ increases we can see qualitatively that the frequency of the reconstructed function increasingly skews towards lower frequencies. 
    We fit polynomial regression to the reconstructed function, picking the optimal polynomial degree $k^*$ via cross-validation on 10 randomly chosen train-test splits.
    As $\sigma_\phi$ increases, the lower frequency content of the decoder function corresponds, as one might expect, to a lower optimal polynomial degree $k^*$. 
    } 
    \label{fig:frequency_illus}
\end{figure*}

Previous analysis in Gaussian spaces has been done for spaces equipped with the standard Gaussian measure (zero mean and unit variance). 
To directly apply the theory of Gaussian spaces to VAEs, we extend the theory of these spaces for standard Gaussian measures to be applicable to \textit{general} Gaussian spaces with arbitrary mean and covariance. 

Using this newly developed theory we consider a basis-expansion of the decoder of a VAE in terms of the eigenfunctions of a general Gaussian space, the \textit{Hermite polynomials}.
This then enables us to see how higher degrees of these basis functions, these polynomials, are implicitly more-strongly penalised as the variance of the underlying Gaussian space increases, leading the decoder of VAE to preferentially learn functions that can be represented as lower degree polynomials. 
As higher-order polynomials are naturally associated with higher frequency content in the Fourier domain, this means that a higher variance Gaussian measure in the latent space necessarily leads to decoders with lower-frequency functional representations.

Further, the encoder, which in our analysis parameterises the Gaussian measure of the latent space, is also affected by changes in the latent measure's variance. 
By studying the Fourier transform of Gaussian measures we can show very simply that larger-variance measures have less high frequency content in the Fourier domain.
This implies that encoders that learn high-variance posteriors have lower-frequency representations. 

To modulate the frequency content of the \textit{individual networks} that constitute the VAE encoder, 
we study the effect of adding Gaussian noise to VAE inputs during training.
This noising operation effectively turns the input space into a Gaussian space on a per-datapoint basis. 
In the context of generative models, this process is reminiscent both of Spread Divergences~\citep{Zhang2020spread} and dequantisation~\citep{Dinh2017, Salimans2017, Ho2019flow}.
As we demonstrate both theoretically and empirically, these noise additions enable finer-grained control over the harmonic content of the encoder mean network. 

Further, we demonstrate that the Kullback-leibler ($\mathrm{KL}$) divergence between the VAE latent prior and variational posterior in the VAE evidence lower bound (ELBO) serves as a single point of action to modulate the frequency content of both encoder and decoder networks; whereby a larger proportional weighting of the  $\mathrm{KL}$ during training induces VAE networks with lower frequency components. 

Finally, using this method of harmonic analysis, we extend Nash’s Poincaré inequality to general Gaussian spaces.
This shows that Gaussian spaces implicitly induce a soft constraint on the Lipschitz constant of their constituent function.
In the VAE setting, we use these results to show that the Lipschitz constant of the VAE decoder decreases as the posterior variance increases.

This result links our work both to recent advances in the Lipschitz penalisation and regularisation of deep generative models~\citep{Adler2018, terjek2020adversarial} and more generally to the adversarial robustness literature where control of the Lipschitz constant improves the robustness of models \citep{gouk2018regularisation, yang2020, hein2017formal, tsuzuku2018lipschitzmargin}.
Our novel theoretical viewpoint on VAEs unifies the two emerging strands in the empirical and theoretical study of the robustness of VAEs to adversarial attack.
The first strand aims to tune the noise of the VAE latent space by up-weighting various regularisation terms ~\citep{Willetts2019a, alex2020theoretical}.
The second aims to directly control the Lipschitz constants of the underlying networks~\citep{barrett2021}.
In our framework we link these two strands: the Gaussian noise of a VAE's latent space affects the Lipschitz constants of the VAE's constituent networks. 

\section{Background}

\paragraph{Variational Autoencoders:}
VAEs~\citep{Rezende2014, Kingma2013}, and the models they have inspired~\citep{Alemi2016,Kumar2020,Chen2018,Willetts2019a}, are deep latent variable models.
Using $\v{x}\in\mathcal{X}$ to denote data and $\v{z}\in\mathcal{Z}$ to denote the latents with associated prior $p(\v{z})$, a VAE simultaneously learns both a forward generative model, $p_\theta(\v{x}|\v{z})$, and an amortised approximate posterior distribution, $q_\phi(\v{z}|\v{x})$ (where $\theta$ and $\phi$ correspond to their respective parameters) which are typically implemented using neural networks~\footnote{
Notation: We use bold letters to denote vectors or matrices and non-bolded letters to denote scalars. Matrices are capitalised. }.
These models are referred to as the decoder and encoder respectively, and a VAE can be thought of as a deep stochastic autoencoder.
Under this autoencoder framework, one typically takes the reconstructions as deterministic, corresponding to the mean of the decoder, namely $g_{\theta}(\v z) := \expect_{p_\theta(\v{x}|\v{z})}[\v x]$, a convention we adopt.

A VAE is trained by maximizing the 
evidence lower bound (ELBO)
$\mathcal{L}\!=\! \expect_{p_{\mathcal{D}}(\v{x})}[\mathcal{L}(\v{x})]$, where
\begin{equation}\label{eq:elbo}
    \ELBO(\v{x})=\expect_{q_\phi(\v{z}|\v{x})} \left[\log p_\theta(\v{x}|\v{z})\right] - \KL(q_\phi(\v{z}|\v{x}) || p(\v{z}))
\end{equation}
and $p_{\mathcal{D}(\v{x})}$ is the empirical data distribution.
The optimisation is carried out using stochastic gradient ascent with Monte Carlo samples to evaluate expectations over $\v z$, typically employing the reparameterisation trick~\citep{Kingma2013}. 
For example, for a Gaussian 
$q_\phi(\v{z}|\v{x})$, we draw samples as \(\v{z} = \v{\mu}_\phi(\v{x}) + \v{\epsilon} \circ \v{\sigma}_\phi(\v{x}), \v{\epsilon} \sim \mathcal{N}(\v{0}, \v{I})\), where $\circ$ is the element-wise product.

Previous work \citep{Camuto2020_2, alex2020theoretical, rezende2018taming} has shown the regularising effects of increasing the noisiness of VAE encodings, by increasing the variance of the variational posterior. 
VAEs are less likely to overfit under these settings and can even be more robust to adversarial attack. 
Our contribution here is to offer a simple framework that allows the study of VAEs from a functional and frequency domain perspective, and gives more precise insights into the polynomial order and spectral properties of encoder and decoder functions (rather than some abstract measure of model complexity \citep{rezende2018taming}) that different encoder variances induce.

\paragraph{Gaussian Spaces:} A \textit{Gaussian Space}, denoted $L_2(\mathbb{R}^n, \gamma)$ is an $L_2$ space, the space of square-integrable functions $f: \mathbb{R}^n \to \mathbb{R}$, equipped with the Gaussian measure $\gamma(x) = \prod_i \mathcal{N}(x_i|0,1)$. This space has an inner product between two functions $f$ and $g$: $\langle f, g \rangle = \expect_{\gamma(x)} \left[f(x)g(x) \right]$. 

On the real line, \textit{the Hermite polynomials} form an \textit{orthogonal basis} for the space $L_2(\mathbb{R}, \gamma)$.
These are the polynomials of degree $k$ that satisfy \citep{janson_1997}
\begin{equation}\label{eq:ortho_hermite}
    \expect_{\gamma(x)}[H_k(x)H_m(x)] = k!\mathbf{1}\{m=k\}.
\end{equation}

These polynomials can also be defined recursively, in a manner that allows a more intuitive understanding of their polynomial nature:
\begin{align}\label{eq:hermite_def}
    H_{k+1}(x) = x H_{k}(x) - k H_{k-1}(x)\\
    H_0(x)=1, H_1(x)=x \nonumber.
\end{align}
We have, for instance, 
\[ H_2(x)= x^2-1, H_3(x)= x^3 - 3x, H_4(x)= x^4 - 6x^2 + 3 .\]

A function $f \in L_2(\mathbb{R}, \gamma)$ in this space can be expressed as a weighted sum of these polynomial functions, where $\hat{f}(k)$ are the \textit{Hermite coefficients}: 
\begin{align}
     f &= \sum_{k \in \mathbb{N}} \frac{1}{k!}\hat{f}(k)H_k \\ \hat{f}(k) = \langle f, H_k \rangle  &=\expect_{\gamma(x)} \left[f(x)H_k(x) \right]. 
\end{align}

\section{Gaussian Spaces and VAEs} 

We want to bring the tools of Gaussian space analysis to bear on VAEs, by viewing the latent space of the VAE as being equipped with a Gaussian measure.
The VAE encoder parameterises an amortised (per-datapoint) posterior Gaussian distribution with mean $\mu_\phi(\v{x})$ and \textit{diagonal variance} $\mathrm{diag}(\sigma^2_\phi(\v{x}))$.
Thus we can view the latent space of a VAE as being equipped with a Gaussian measure that varies on a \textit{per-datapoint basis}.
This measure is a multivariate general Gaussian measure with mean $\mu_\phi(\v{x})$ and \textit{diagonal variance} $\mathrm{diag}(\sigma^2_\phi(\v{x}))$. 
The encoder in some sense `indexes' over the range of possible Gaussian measures in a data-dependent manner. 
The decoder, which acts on this latent space, would then be a member of some Gaussian space (the measure for which depends on $\v{x}$). 

As we were unable to find a comprehensive resource for spaces equipped with non-standard Gaussian measures we must first derive results for Gaussian spaces equipped with a \textit{general} Gaussian measure (with mean $\mu$ and variance $\sigma^2$) before we apply Gaussian space analysis to VAEs. 
We start with results for functions acting on a univariate space, $f:\mathbb{R} \to \mathbb{R}$. 

\begin{proposition} \label{prop:general_gauss_space}
If $x\sim\mathcal{N}(\mu,\sigma^2)$, then we have that $\hat{x} = (x-\mu)/\sigma \sim \mathcal{N}(0,1)$ and the Hermite polynomials are the polynomials of degree $k$ that satisfy 
\[
\expect_{\gamma(\hat{x})}[H_k(\hat{x})H_m(\hat{x})] = k!\mathbf{1}\{m=k\}, 
\]
where $\gamma$ is the standard Gaussian measure. 
The family $\left\{\frac{1}{\sqrt{k!}}H_k(\hat{x}): k \geq 0 \right\}$ is then an orthonormal basis for $L_2(\mathbb{R}, \mathcal{N}(\mu,\sigma^2))$. 
\end{proposition}
For the proof see Appendix~\ref{app:general_gauss_space}\footnote{\label{noteproof} All proofs are presented in Appendix~\ref{app:proofs}}. As such the polynomials $H_n(\hat{x})$ are the orthogonal polynomials for the $\mathcal{N}(\mu, \sigma^2)$ measure. 
We can calculate each of these polynomials by substituting $x$ for $\hat{x}=(x-\mu)/\sigma$ in Equation~\eqref{eq:hermite_def}.
In higher dimensions ($\mathbb{R}^n$), for a function $f:\mathbb{R}^n \to \mathbb{R}$, we assume that we have a \textit{diagonal} standard deviation $\v{S}$, as this form of the standard deviation is most directly applicable to how VAEs are used in practice.

We denote the multivariate Gaussian measure as $\mathcal{N}(\v{\mu}, \v{S}^2)$ with covariance matrix populated by the element-wise square of $\v{S}$. 
In this case the basis can be expressed using a multi-index $\v{\alpha} \in \mathbb{N}^n$ (the sets of size $n$ of non-negative integers),
where
\begin{align*}
    |\v{\alpha}| &= \sum_{i=1}^n\alpha_i,\qquad \v{v}^{\v{\alpha}} = \prod_{i=1}^n v_i^{\alpha_i} \\ \qquad \v{\alpha}! &= \prod_{i=1}^n \alpha_i!, \qquad g^{(\v{\alpha})}(\v{x}) = \frac{\partial^{|\v{\alpha}|}g}{\partial x^{\alpha_1}_1\dots\partial x^{\alpha_n}_n}.
\end{align*}
The $\v{\alpha}^{\mathrm{th}}$ multivariate Hermite polynomial (denoted $\mathcal{H}_{\v{\alpha}}$) for the measure $\mathcal{N}(\v{\mu}, \v{S}^2)$ can then be expressed as the product of univariate polynomials indexed by $\v{\alpha}$:
\[\mathcal{H}_{\v{\alpha}}((\v{x}-\v{\mu})\v{S}^{-1})=\mathcal{H}_{\v{\alpha}}(\v{\hat{x}})=\prod_i H_{\alpha_i}((x_i-\mu_i)/\sigma_i),\]
This stems from the fact that each $H_{\alpha_i}$ forms a basis in the univariate case, and that we assume a diagonal covariance, meaning that we can obtain the basis for $\mathbb{R}^n$ simply by tensorisation $H_{\alpha_1} \otimes \cdots \otimes H_{\alpha_n}$.
For the sake of completeness, in Appendix~\ref{app:hermite_full_cov} we derive the Hermite polynomials for a Gaussian space with a \textit{full covariance matrix}. 

Because the set of $\mathcal{H}_{\v{\alpha}}$ form an orthogonal basis, we can express functions $f \in L_2(\mathbb{R}^n, \mathcal{N}(\v{\mu}, \v{S}^2))$ as a weighted sum of these functions: 
\begin{align}
    f &= \sum_{\v{\alpha} \in \mathbb{N}^n} \frac{1}{\v{\alpha}!} \hat{f}(\v{\alpha})\mathcal{H}_{\v{\alpha}} \\
     \hat{f}(\v{\alpha}) &=\expect_{\gamma(\v{\hat{x}})} \left[f(\v{x})\mathcal{H}_{\v{\alpha}}(\v{\hat{x}}) \right] \nonumber \\ \gamma(\v{\hat{x}}) &= \mathcal{N}(\v{x}|\v{\mu},  \v{S}^2). \nonumber
\end{align}

Using these results,  we can express the Hermite coefficients for a general Gaussian space in terms of the underlying measure's standard deviation $\v{S}$ and the derivatives of a function $f$ in the space. 
\begin{proposition} 
\label{prop:hermite_coeffs}
Assume we have a function $f$ in Gaussian space with diagonal covariance, $f \in L_2(\mathbb{R}^n, \mathcal{N}(\v{\mu}, \v{S}^2))$. Further assume that $f$ is in $C^\infty$, the class of infinitely differentiable functions.
For $\v x\sim\mathcal{N}(\v{\mu},\v{S}^2)$, then we have that $\v{\hat{x}} = (\v{x}-\v{\mu})\v{S}^{-1} \sim \mathcal{N}(\v{0},\v{I})$ and the Hermite coefficients can be expressed as
\begin{align*}
    \hat{f}(\v{\alpha}) &=  (\mathrm{diag}(\v{S}))^{\v{\alpha}}  \expect_{\gamma(\v{\hat{x}})}\left[ f^{(\v{\alpha})}(\v{x})\right], \gamma(\v{\hat{x}}) = \mathcal{N}(\v{x}|\v{\mu}, \v{S}^2).
\end{align*}
\end{proposition}

We can now express the variance of a function $L_2(\mathbb{R}^n, \mathcal{N}(\v{\mu}, \v{S}^2))$ as a sum of function derivatives, which in turns allows us to connect this variance to the Fourier domain.
\begin{theorem} \label{thm:var_fourier} 
Assume we have a function $f$ in Gaussian space with diagonal covariance, $f \in L_2(\mathbb{R}^n, \mathcal{N}(\v{\mu}, \v{S}^2))$.
Further assume that $f$ is in $C^\infty$, the class of infinitely differentiable functions and is $L_2$ integrable with respect to the Lebesgue measure.
For $\v{x}\sim\mathcal{N}(\v{\mu},\v{S}^2)$, then we have that $\v{\hat{x}} = (\v{x}-\v{\mu})\v{S}^{-1} \sim \mathcal{N}(\v{0},\v{I})$ such that $\gamma(\v{\hat{x}}) = \mathcal{N}(\v{x}|\v{\mu}, \v{S}^2)$. 
The variance of $f$ can be expressed as 
\begin{align*}
    \mathrm{Var}(f) &=  \sum_{|\v{\alpha}| \geq 1 } \frac{(\mathrm{diag}(\v{S}))^{2\v{\alpha}}}{\v{\alpha}!} \left|\expect_{\gamma(\hat{x})}\left[ f^{(\v{\alpha})}(\v{x})\right]\right|^2 \\ &= \sum_{|\v{\alpha}| \geq 1 } \frac{(\mathrm{diag}(\v{S}))^{2\v{\alpha}}}{\v{\alpha}!} 
    \left|\int_{\mathbb{R}^n} (\v{\omega})^{\v{\alpha}}\mathcal{F}(\v{\omega})\overline{\mathcal{P}(\v{\omega})}d\v{\omega}\right|^2, 
\end{align*}
where $\mathcal{P}$ is the Fourier transform of the Gaussian measure $\mathcal{N}(\v{\mu},\v{S}^2)$ given by 
$\mathcal{P}(\v{\omega})= \mathrm{det}(\v{S})\mathcal{G}(\v{\omega}\v{S})e^{-i\v{\omega}\v{\mu}\v{S}^{-1}}$ (where $\mathcal{G}$ is the Fourier transform of the standard Gaussian measure $\gamma$) and $\mathcal{F}$ is the Fourier transform of $f$, and $\v{\alpha} \in \mathbb{N}^n$.
\end{theorem}

We also note that even if a function is \textit{not} infinitely differentiable, we can also express $\mathrm{Var}(f)$ as a weighted sum of the Hermite coefficients
(see the Proof of Theorem~\ref{thm:var_fourier}):
\begin{equation}
    \mathrm{Var}(f) =  \sum_{|\v{\alpha}| \geq 1 } \frac{1}{\v{\alpha}!} \left|\hat{f}(\v{\alpha})\right|^2.
\end{equation}
If we now assume that $f$ is infinitely differentiable, then we obtain the intuitive result that high-frequency functions correspond to larger Hermite coefficients associated with higher degree Hermite polynomials. 
This can be deduced by combining Proposition~\ref{prop:hermite_coeffs} and Theorem~\ref{thm:var_fourier}:
\begin{equation} \label{eq:coeffs_fourier}
    \left|\hat{f}(\v{\alpha})\right|^2 = \left|\int_{\mathbb{R}^n} (\mathrm{diag}(\v{S}))^{\v{\alpha}} (\v{\omega})^{\v{\alpha}}\mathcal{F}(\v{\omega})\overline{\mathcal{P}(\v{\omega})}d\v{\omega}\right|^2.
\end{equation}
As the variance of the underlying Gaussian measure increases, larger degree polynomials contribute more heavily to $\mathrm{Var}(f)$. 
This is captured in the terms $(\mathrm{diag}(\v{S}))^{2\v{\alpha}}$ of Equation~\eqref{eq:coeffs_fourier}, meaning that increases in $|(\mathrm{diag}(\v{S}))|$ disproportionately increase large $\v{\alpha}$ terms, i.e the coefficients associated with higher degree polynomial terms. 
More succinctly, Gaussian spaces with large variances naturally induce larger Hermite coefficients associated with higher degree polynomials, which themselves are associated with higher frequency components in the Fourier domain.

\paragraph{Lipschitzness}
Finally we redevelop Nash’s Poincaré inequality for general Gaussian spaces to show that the variance of a function $f \in L_2(\mathbb{R}^n, \mathcal{N}(\v{\mu}, \v{S}^2))$, assuming that this function is Lipschitz continuous, can be upper-bounded by using its Lipschitz constant.
\begin{proposition}\label{prop:lipschitz_hermite}
Let $f \in L_2(\mathbb{R}^n, \mathcal{N}(\v{\mu}, \v{S}^2))$ be a function that is Lipschitz continuous with Lipschitz constant $L$.
Further assume that $f^{(|\v{\alpha}|=1)} \in L_2(\mathbb{R}^n, \mathcal{N}(\v{\mu}, \v{S}^2))$, we have
\[\mathrm{Var}(f)  \leq L^2\|\v{S}\|_2^2.\]
\end{proposition}
This bound is relatively tight in that for a measure $\mathcal{N}(\v{0},\v{I})$ and a function $f(\v{x}) = \frac{1}{n} \sum x_i$ the bound becomes an equality: $L=n^{-1}$ and  $\mathrm{Var}(f) = n^{-2}$.
We can now use these results for a general Gaussian space to study VAE decoders.

\subsection{Gaussian Spaces and VAE decoders} 
\label{sec:decoder_freq}

The amortised posterior distributions of a VAE are typically chosen as Gaussians, where, to allow for the backpropagation of gradients, samples are reparameterised as
\begin{equation} \label{eq:reparam}
    \v{z} = \v{\mu}_{\phi}(\v{x}) + \v{\sigma}_{\phi}(\v{x}) \circ \v{\epsilon}, \qquad \v{\epsilon} \sim \mathcal{N}(\v{0},\v{I}).
\end{equation}
This sample is then fed into the decoder to calculate the likelihood term of the ELBO (see Eq~\eqref{eq:elbo}). 
Implicit in this arrangement is that the VAE decoder operates on samples $\v{z}$ in latent space $\mathcal{Z}$ equipped with a Gaussian measure, captured by the  $\v{\epsilon}$ term in Eq~\eqref{eq:reparam}. 

Assuming the function parameterised by the decoder $g$, for a given point $\v x$, is $L_2$ integrable with respect to the general Gaussian measure, i.e  $g \in L_2(\mathbb{R}^n, \mathcal{N}(\v{\mu}_{\phi}(\v{x}), \v{\sigma}_{\phi}(\v{x})))$ (this is reasonable as we are effectively training the decoder to be in this space) then we can directly apply the theory of Gaussian spaces to the VAE decoder. 
Assuming we have a Gaussian likelihood with a fixed standard deviation $\sigma_\theta$ across dimensions, we express the likelihood for a single point using a bias-variance decomposition, which gives the sum of an error term captured by the bias, and a regularisation term captured by the variance:
\begin{align}
    &\frac{1}{\sigma_\theta^2}\expect_{q_\phi(\v{z}|\v{x})} \left[(g_{\theta}(\v{z}) - \v{x})^2 \right] \nonumber \\ &= \frac{1}{\sigma_\theta^2} \left( \left(\mathrm{Bias}_{q_\phi(\v{z}|\v{x})}\left( g_{\theta}(\v{z}) \right)\right)^2 + \mathrm{Var}_{q_\phi(\v{z}|\v{x})}(g_{\theta}(\v{z}))\right) \\
   &\mathrm{Bias}_{q_\phi(\v{z}|\v{x})}\left( g_{\theta}(\v{z}) \right) = \expect_{q_\phi(\v{z}|\v{x})}\left[ g_{\theta}(\v{z}) \right] -\v{x} \nonumber \\
   &\mathrm{Var}_{q_\phi(\v{z}|\v{x})}\left( g_{\theta}(\v{z}) \right) \nonumber \\ & \qquad\qquad= \expect_{q_\phi(\v{z}|\v{x})}\left[\left( g_{\theta}(\v{z}) - {\expect_{q_\phi(\v{z}|\v{x})}}\left[ g_{\theta}(\v{z})\right] \right)^2 \right]. \nonumber
\end{align}

We can directly apply Theorem~\ref{thm:var_fourier} to the decoder function if we assume that the decoder function is infinitely differentiable.
This holds for networks with the $\mathrm{Sigmoid}$ activation function for instance \citep{Hornik1991}. 
We can then swap out $\mathrm{diag}(\v{S})$ for the encoder standard deviation $\v{\sigma}_{\phi}(\v{x})$ and repeatedly apply the Theorem to each output $g_{\theta,i}(\v{z}), i \in 1,\dots, d$ of the decoder.
For an output $i$ of the decoder and $\v{\alpha} \in \mathbb{N}^n$, we have that 
\begin{align}
     &\mathrm{Var}_{q_\phi(\v{z}|\v{x})}(g_{\theta,i}) =  \sum_{|\v{\alpha}| \geq 1 } \frac{(\v{\sigma}_{\phi}(\v{x}))^{2\v{\alpha}}}{\v{\alpha}!} \left|g_{\theta,i}^{(\v{\alpha})}(\v z)\right|^2 \nonumber \\ &= 
     \sum_{|\v{\alpha}| \geq 1 } \frac{(\v{\sigma}_{\phi}(\v{x}))^{2\v{\alpha}}}{\v{\alpha}!} 
    \left|\int_{\mathbb{R}^n} (\v{\omega})^{\v{\alpha}}\mathcal{F}_{\theta,i}(\v{\omega})\overline{\mathcal{P}_{\phi}(\v{\omega})}d\v{\omega}\right|^2.
\end{align}

$\mathcal{P}_{\phi}$ is the Fourier transform of the Gaussian measure $\mathcal{N}(\mu_\phi(\v{x}),\sigma^2_\phi(\v{x}))$ and $\mathcal{F}_{\theta,i}$ is the Fourier transform of $g_{\theta,i}$.

As $\v{\sigma}_{\phi}(\v{x})$ increases, particularly for $\v{\sigma}_{\phi}(\v{x}) \geq 1$, larger $\v{\alpha}$ terms in the sum, and large frequencies to the power $\v{\alpha}$ will begin to be disproportionately penalised in the $\mathrm{Var}$ term of the bias variance decomposition of the VAE likelihood. 
Thus larger $\v{\sigma}_{\phi}(\v{x})$ will result in a larger penalisation of the high-frequency components of the decoder function.
We note that this penalisation is modulated on a \textit{per-datapoint basis} by $\v{\sigma}_{\phi}(\v{x})$, but that VAEs which on average have larger encoder variances over training inputs will learn lower frequency decoders.
Because of the link between the decoder variance, the Hermite polynomials, and the frequency domain (see Equation~\eqref{eq:coeffs_fourier}), the lower frequency function learned by the decoder, for each $\v{z}|\v{x}$, \textit{also} corresponds to a function that can be described as a lower degree polynomial.

\begin{figure}[t!]
  \begin{center}
    \includegraphics[width=0.29\textwidth]{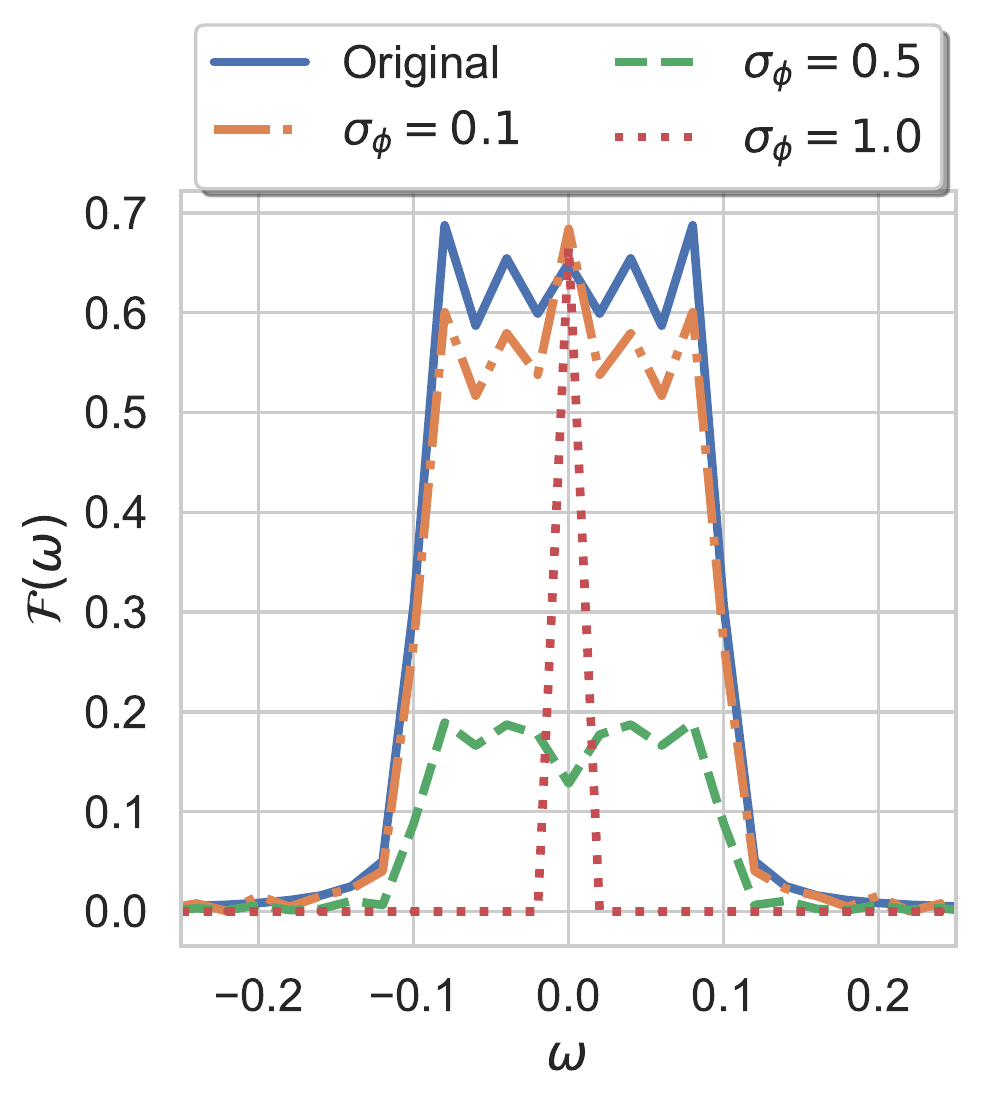}
    \includegraphics[width=0.29\textwidth]{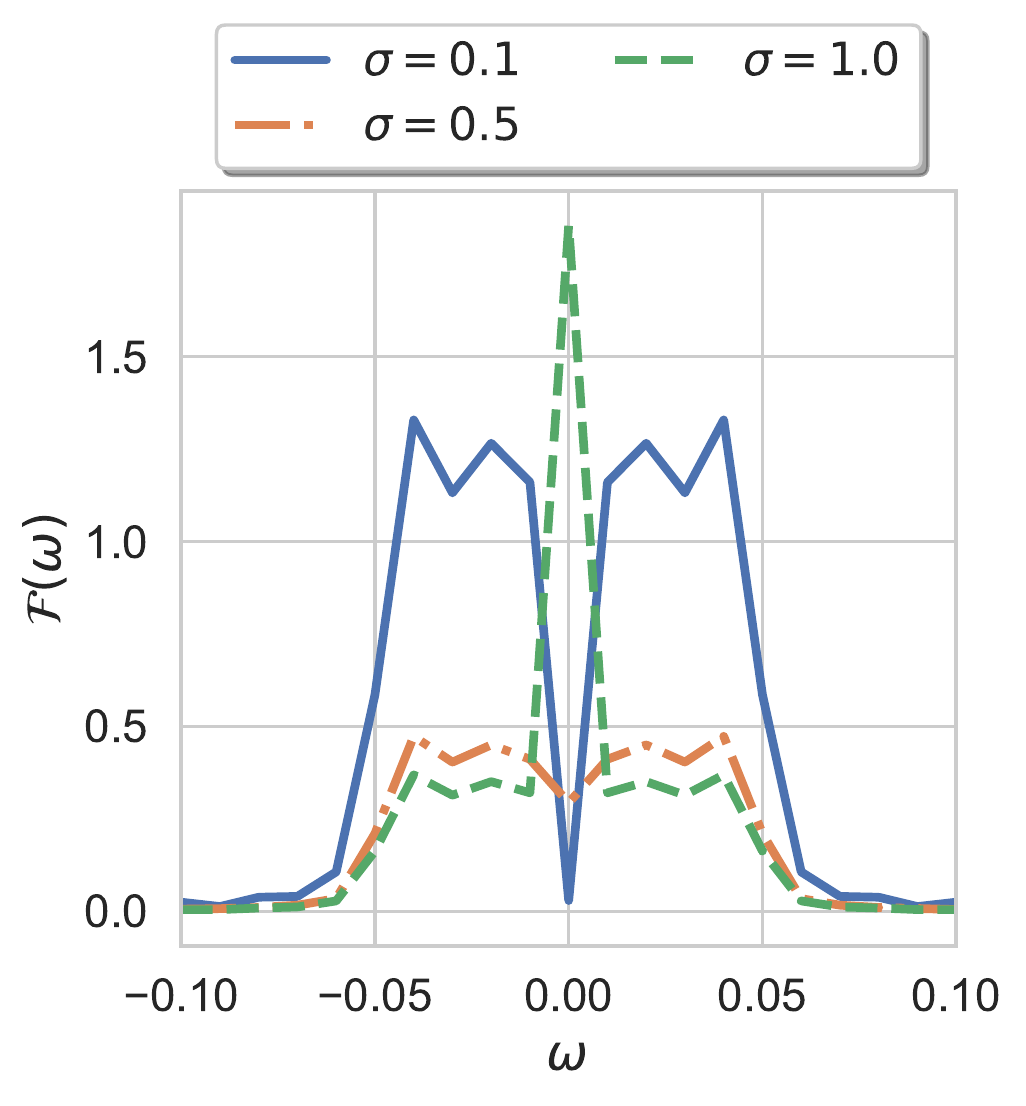}
  \end{center}
  \caption{
  [top] FFT of the VAE reconstructions in Figure~\ref{fig:frequency_illus}. For larger $\v{\sigma}_{\phi}$, the spectrum loses high frequencies.
  [bottom] FFT of $\mu_\phi$ of VAEs trained on $\mathrm{sinc}(5t) + \mathcal{N}(0,\sigma^2\v{I})$. For larger $\sigma$ the spectrum loses high frequency content. 
 }
\label{fig:fftsinc}
  \label{fig:frequency_illus_latent}
\end{figure}

To demonstrate these findings, in Figure~\ref{fig:frequency_illus} we use a fixed encoder variance $\sigma_{\phi}^2$, uniform across dimensions, on VAEs, with $\mathrm{Sigmoid}$ activation function in their networks, trained on data from the $\mathrm{sinc}(5t), t\in[-1,1]$ function. 
As $\sigma_{\phi}$ increases, we can qualitatively ascertain that the decoder function loses higher frequency components and that the optimal polynomial to describe the decoder function (approximated by polynomial regression) \textit{decreases} in its degree. 
Quantitatively measuring the Fourier transform of the decoder here is simple, we collect the reconstruction of all inputs, ordering by increasing $t$. We then take the fast Fourier transform (FFT) of this aggregate reconstructed function. 
We show this in Figure~\ref{fig:fftsinc} where as $\sigma_{\phi}$ increases, the decoder learns a lower frequency Fourier representation. 

\begin{figure*}[t!]
\centering
    \includegraphics[width=0.25\textwidth]{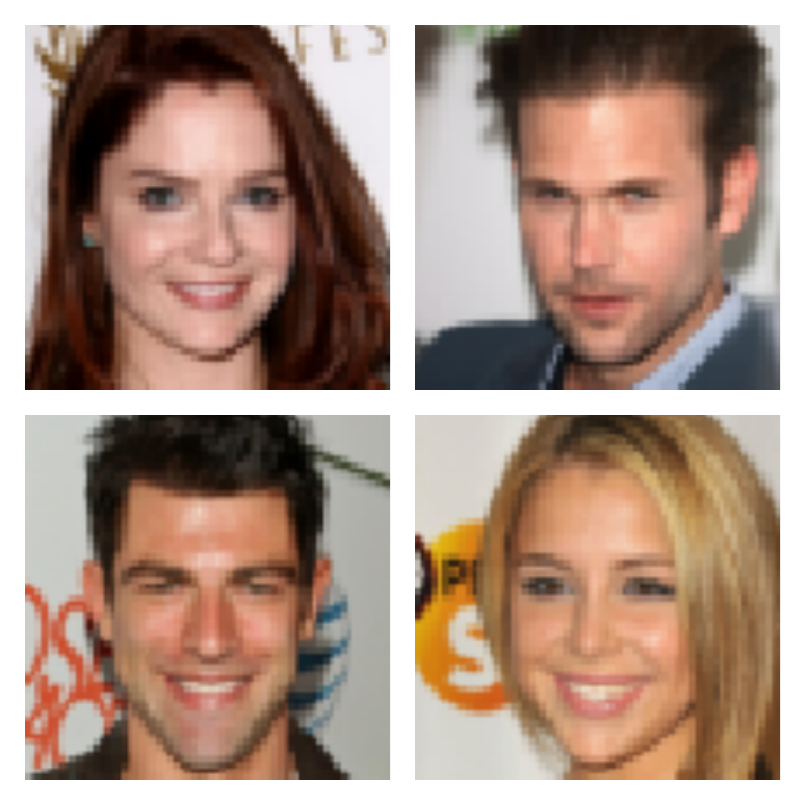}
    \hspace{1.5em}
     \includegraphics[width=0.25\textwidth]{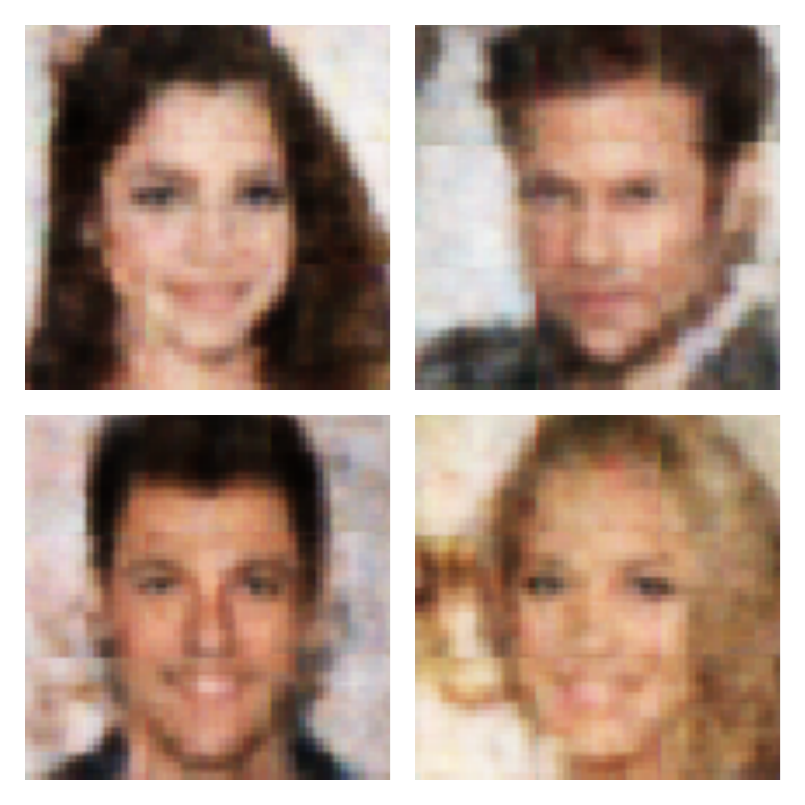}
     \hspace{1.5em}
     \includegraphics[width=0.25\textwidth]{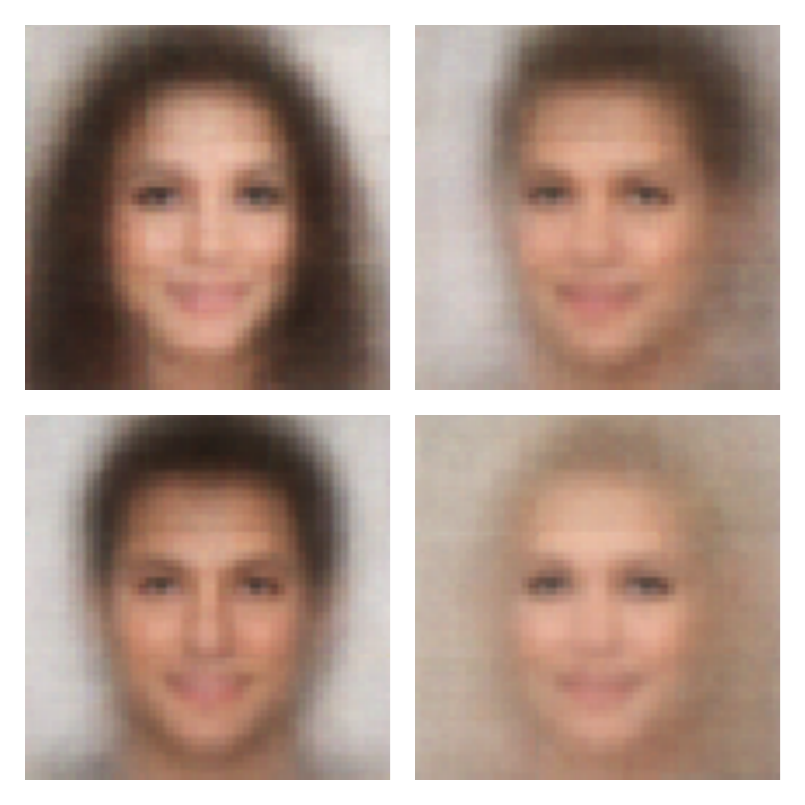}\\
        \subfloat[][Inputs]{
    \includegraphics[width=0.25\textwidth]{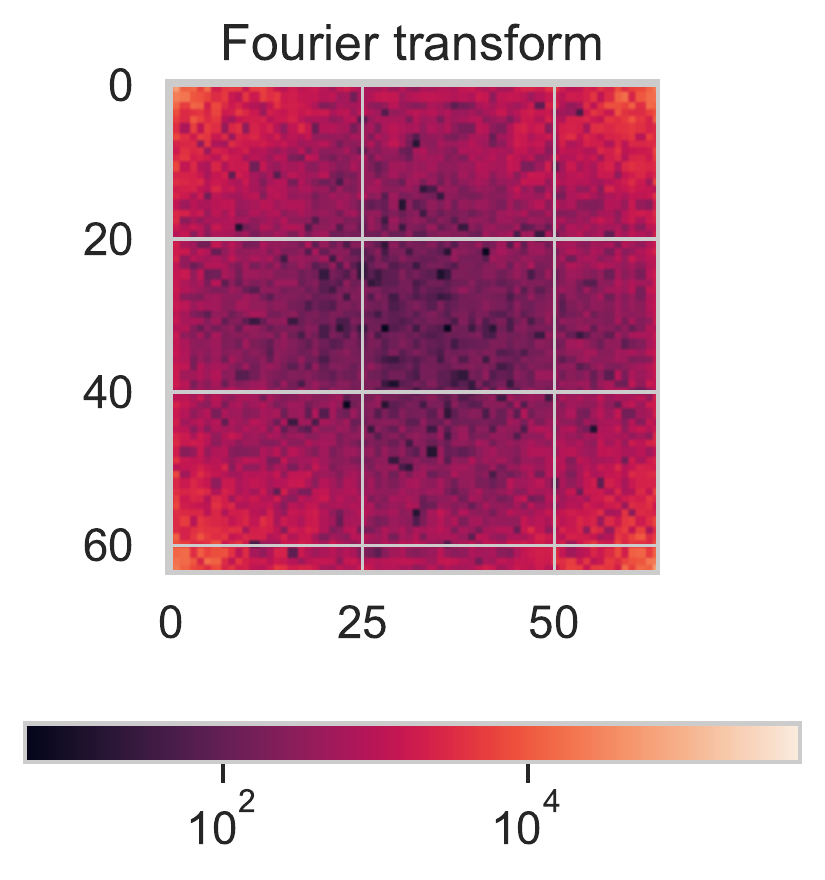}}
    \hspace{1.5em}
            \subfloat[][$\sigma_\phi=1$]{
     \includegraphics[width=0.25\textwidth]{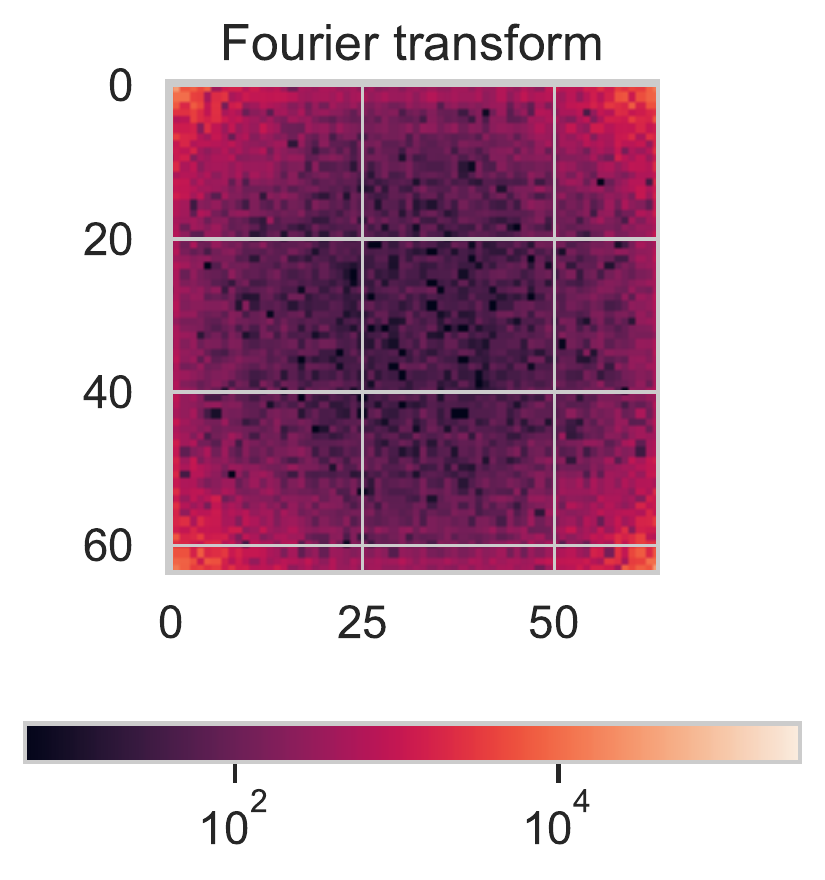}}
      \hspace{1.5em}
            \subfloat[][$\sigma_\phi=2$]{
    \includegraphics[width=0.25\textwidth]{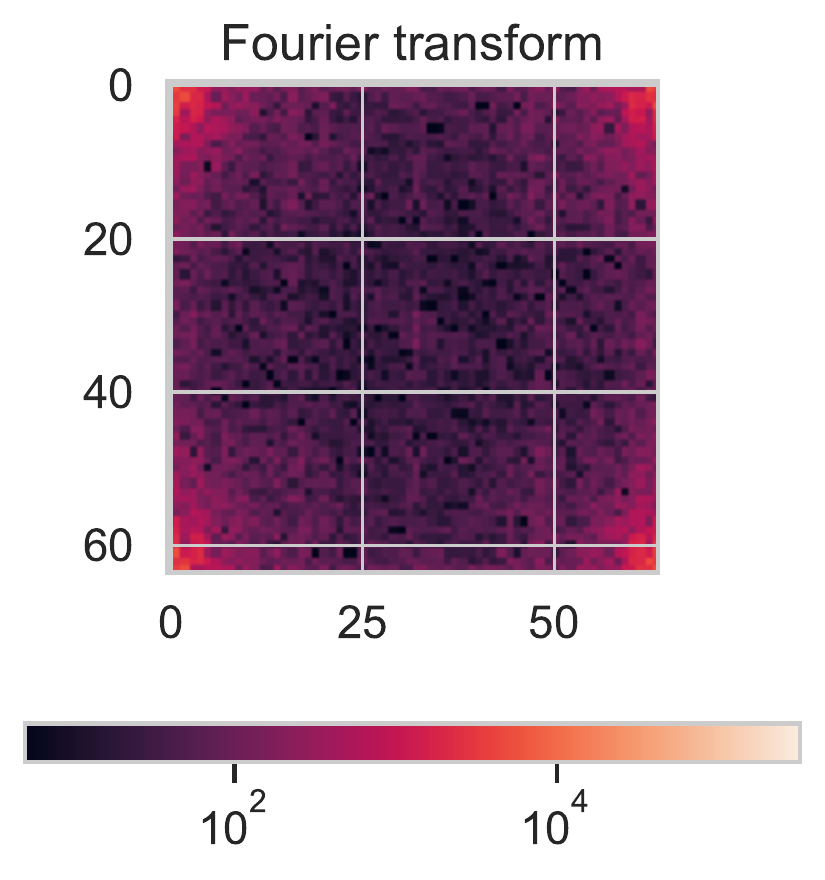}}
    \\
     \includegraphics[width=0.5\textwidth]{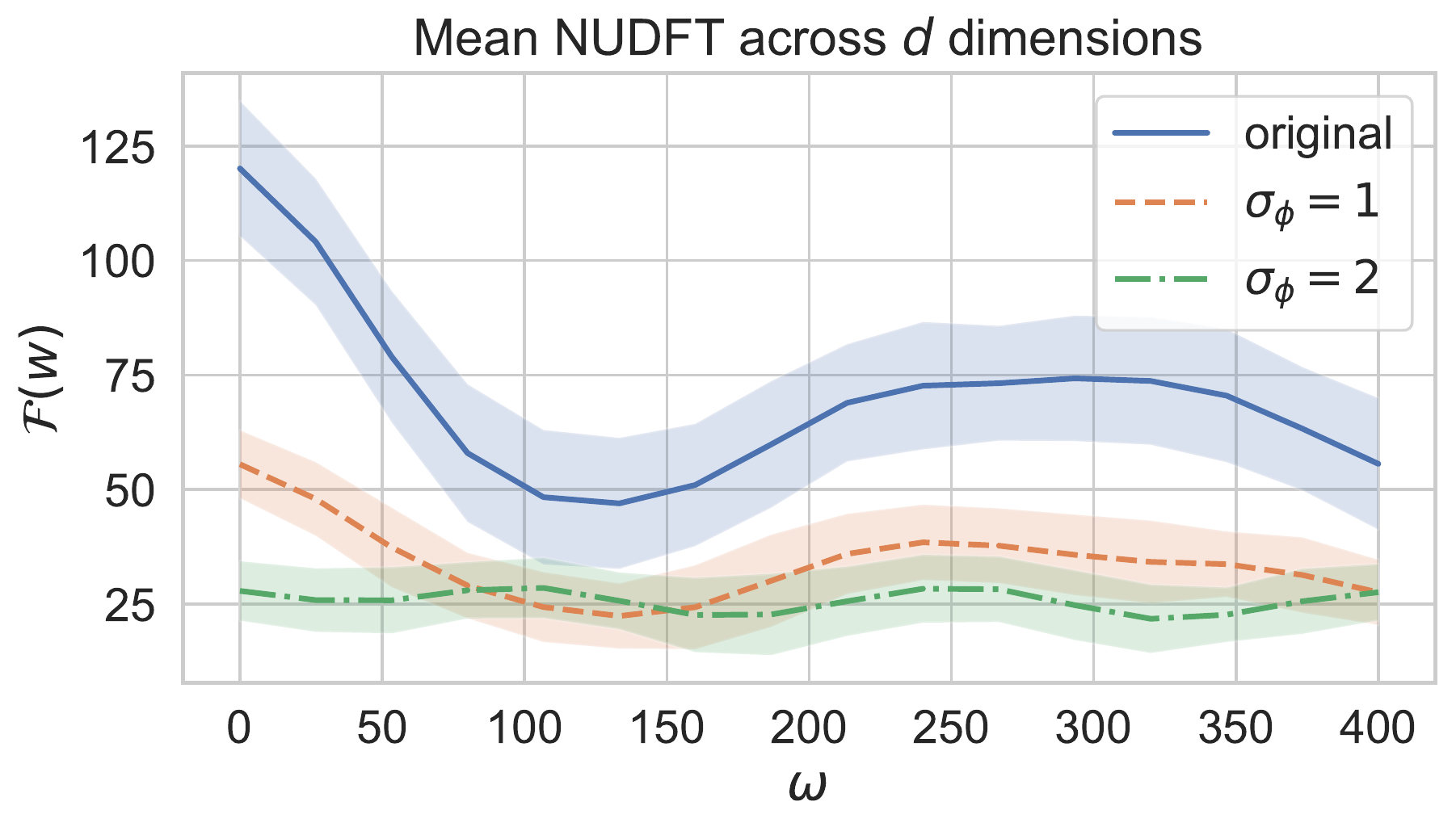}
    \caption{We train convolutional VAEs (see Appendix~\ref{app:architecture} for network details), with $\mathrm{ReLU}$ activation functions in their networks, on the CelebA dataset, with a 64 dimension latent space $\mathcal{Z}$. As before we fix the encoder $\sigma_\phi$. 
    \textbf{[first + second rows]} We show 4 images and the 2D-FFT of the image in the upper right quadrant for a) the training images used b) VAE reconstructions of the 4 images when $\sigma_\phi=1$ c) VAE reconstructions for $\sigma_\phi=2$. 
    As $\sigma_\phi$ increases mid-high level frequencies are increasingly dampened relative to the original image.
    \textbf{[bottom row]} Positive components of the mean 1D-NUDFT of the $d$ dimensions of the output of these models (calculated across 2000 images for each of the $d=12288$ dimensions). 
    Shading corresponds to the std. dev. over $d$ dimensions.
    } 
    \label{fig:frequency_illus_celeba}
\end{figure*}

Though our theoretical results hold for classes of infinitely differentiable neural networks, we empirically confirm they still hold for more complex neural network architectures with $\mathrm{ReLU}$ activation functions, trained on large multivariate datasets.
In Figure~\ref{fig:frequency_illus_celeba} we show that for a fixed encoder variance $\sigma^2_\phi$, as $\sigma_\phi$ increases reconstructed CelebA images from a convolutional VAE become more similar and start to lose diversity -- the images lose mid to high level frequencies as measured by a 2D-FFT -- suggesting that the decoder learns a lower frequency function. 
To support this, we plot the mean 1D-Non-uniform discrete Fourier transform (NUDFT) \citep{nudft} across the $d$ dimensions of the output, which shows that models trained with larger $\sigma_\phi$ learn functions that on average are of much lower frequency content than smaller $\sigma_\phi$ models.
This demonstrates that the multi-dimensional output decreases in its high-frequency content over the $d$ output dimensions, as predicted by our theory. 
In Figure~\ref{fig:frequency_illus_multi} of the Appendix we show that this also holds for fully-connected VAEs trained on multivariate sinusoids.

\subsection{Gaussian Spaces and VAE encoders}\label{sec:encoder_freq}
Here we use the analysis developed in previous section to analyze the Harmonic content of the VAE encoder. 
In the typical VAE setup, we don't usually consider the case of Gaussian noise on inputs and we make no assumptions as to the underlying measure of the input space; such that we cannot directly apply the previous Gaussian space analysis to the VAE encoder. 

The perspective we've established, however, of the latent space as a Euclidean space equipped with a general Gaussian measure, and the perspective of the decoder as belonging to Gaussian space equipped with this same measure, can inform our analysis. 
The encoder of the VAE effectively parameterises the latent space measure for each input $\v{x}$ such that the function the encoder is learning is the measure  $\mathcal{N}(\v{\mu}_{\phi}(\v{x}), \v{\sigma}_{\phi}(\v{x}))$. 
In Theorem~\ref{thm:var_fourier} we have already established that the Fourier transform of this measure is given by \[\mathcal{P}(\v{\omega}) = \mathrm{det}(\v{\Sigma}_{\phi}(\v{x}))\mathcal{G}(\v{\omega}\v{\Sigma}_{\phi}(\v{x}))e^{-i\v{\omega}\v{\mu}_{\phi}(\v{x})\v{\Sigma}^{-1}_{\phi}(\v{x})},\]
where $\v{\Sigma}_{\phi}(\v{x})$ is a diagonal matrix with the elements of $\v{\sigma}_{\phi}(\v{x})$ on its diagonal. 
Much like the decoder, as the encoder variance decreases, the function parameterised by the encoder increases in amplitude in its high-frequency components.  
However, here the frequency content of the encoder function is \textit{intrinsic} to the Gaussian measure, whereas the frequency content of the decoder is a result of the relative weighting of the decoder variance in the VAE likelihood, and as such is enforced by optimisation.
Note that $e^{-i\v{\omega}\v{\mu}_{\phi}(\v{x})\v{\Sigma}^{-1}_{\phi}(\v{x})}$ only modulates the phase of the Fourier transform and does not alter the amplitude of its frequency content. 
As such the mean does not affect the frequency spectrum of the posterior Gaussian. 
In Figure~\ref{fig:gaussian_ft} of the Appendix we demonstrate that for a univariate Gaussian, decreasing the variance increases the magnitude of high-frequency components in the Fourier domain, whereas altering the mean has no such effect. 

This perspective gives us an idea of how to alter the harmonic content of the measure parameterised by the encoder \textit{overall}. 
If instead we want to modulate the frequency content of the \textit{individual networks} that are used to parameterise the Gaussian measure we need a different perspective. 
In the next section we show that using the theory developed previously, adding Gaussian noise to the input data offers a simple and effective method for modulating the frequency content of the encoder mean. 

\subsubsection{Noisy inputs}
\label{sec:noisy_inputs}

Let us now assume that we add Gaussian noise to the data $\v{x}$ such that our input for each point $\v{x}$ is a Gaussian space with a Gaussian measure with mean $\v{x}$, variance $\sigma^2$: 
\[\tilde{\v{x}} = \v{x} + \sigma\v{\nu},  \v{\nu} \sim \mathcal{N}(\v{0},\v{I}).\]
Under this noisy input, the expectation of the $\KL$ between the amortised posterior distribution and the unit Gaussian prior can be written as
\begin{align}
    &\expect_{\gamma(\v{\nu})}\left[\mathrm{KL}(q_{\phi}(\v{z}|\tilde{\v{x}})\| p(\v{z}))\right] \nonumber \\ &= \frac{1}{2}\expect_{\gamma(\v{\nu})}\biggl[\log\left( \frac{1}{\prod_{i=1}^n\sigma_{\phi,i}(\tilde{\v{x}})}\right) \nonumber \\ & \qquad - n + \sum_{i=1}^n \sigma_{\phi,i}(\tilde{\v{x}}) + \left\|\v{\mu}_{\phi}(\tilde{\v{x}})\right\|_2^2 \biggr].
\end{align}

We now have a Gaussian space variance $\mathrm{Var}(\v{\mu}_{\phi}) = \expect_{\gamma(\v{\nu})}\left[\left\|\v{\mu}_{\phi}(\tilde{\v{x}})\right\|_2^2 \right]$ penalised at each iteration.
As before for the decoder we can directly apply Theorem~\ref{thm:var_fourier} to the encoder mean function, by swapping out $\mathrm{diag}(\v{S})$ for the data standard deviation $\sigma\v{I}$ and repeatedly applying the Theorem to each output $\mu_i(\tilde{\v{x}}), i \in 1,\dots, n$ of the encoder.
We can expect two things to happen as $\sigma$ increases.
The frequency content of the encoder mean should decrease, and generally the function for each $\v{x}$ can be described using an increasingly lower degree polynomial. 
Thus we can add Gaussian noise to a VAE's inputs to modulate the frequency content of the encoder mean function. 
We demonstrate these findings in Figure~\ref{fig:frequency_illus_latent}.
Note that though our theory is not directly applicable to the encoder variance network, if we fix $\sigma_{\phi}$  so as to modulate the variance of the decoder, then we have full control over the variance of the encoder by adding noise to the VAE inputs.

\subsection{The KL and frequency content} 

In previous experiments we fixed the variance of the encoder for analytical purposes. In practice however we often want to learn this variance, and as such, need tools by which we can modify the frequency content of the encoder and decoder whilst learning the encoder variance. 
As we now show, the frequency content of both the encoder mean and decoder networks can be controlled by altering the weighting of $\mathrm{KL}$ in the ELBO.

Consider the $\beta$-VAE setting \citep{Higgins2017betaVAELB}, whereby the $\mathrm{KL}$ component of the ELBO is pre-multiplied by a factor $\beta$. 
The strength of this penalisation changes the optimal encoder variance $\sigma_{\phi}(x)$: as $\beta$ increases we can expect $\sigma_{\phi}(x)$ to tend to $1$ -- see Theorem 2 of \citet{Camuto2020_2} for a formal statement of this. 
This then means that we can modulate the scale of the latent noise experienced by the model during training, where $\sigma_{\phi}(x)$ is learnt and no longer fixed as in previous experiments. Because increasing $\beta$ tends to increase $\sigma_{\phi}(x)$, our theory would predict that gradual increases in $\beta$ would be accompanied by the decoder losing higher-frequency content in the Fourier domain.

\begin{figure}[t!]
  \begin{center}
    \includegraphics[width=0.29\textwidth]{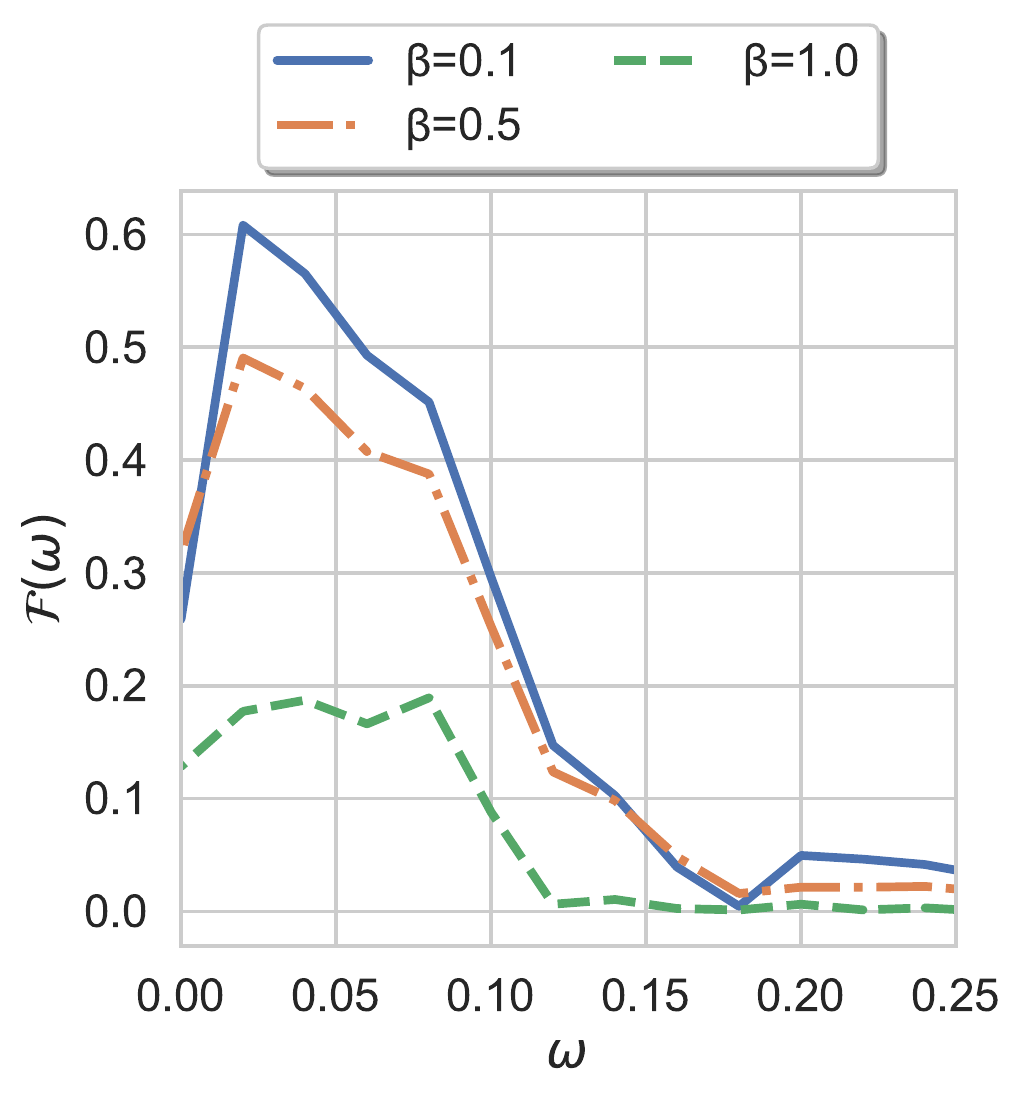}
    \includegraphics[width=0.29\textwidth]{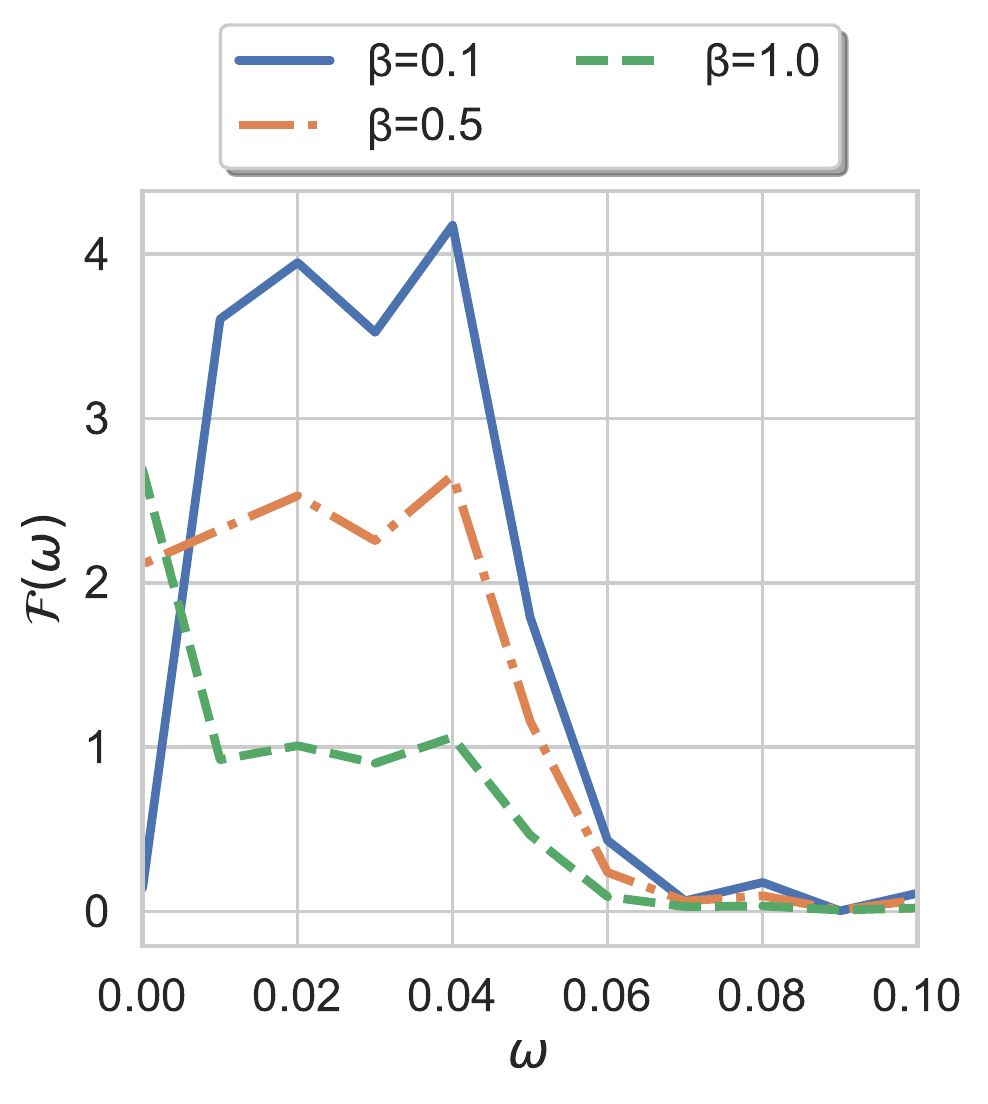}
  \end{center}
  \caption{
  Positive components of Figure \ref{fig:frequency_illus_latent} with variable $\beta$ penalisation on the KL term of the ELBO. As $\beta$ increases higher frequencies are increasingly dampened for the decoder [top] and encoder [bottom]}
  \label{fig:fftsincbeta}
\end{figure}

We also note that the variance of the encoder mean network $\mathrm{Var}(\v{\mu}_{\phi}) = \expect_{\gamma(\v{\nu})}\left[\left\|\v{\mu}_{\phi}(\tilde{\v{x}})\right\|_2^2 \right]$ experiences a greater penalisation as the weight $\beta$
increases, as seen in the decomposition of the KL in section~\ref{sec:noisy_inputs}.
This means that we expect larger $\mathrm{KL}$ weightings to induce encoder mean functions with lower frequency content (when the input space is a Gauss space).
See Figure \ref{fig:fftsincbeta} for a demonstration of both these claims.

\subsection{VAE Lipschitzness} 
We know that by Proposition~\ref{prop:lipschitz_hermite} the variance of a function (that is once differentiable) provides a relatively tight lower bound on the Lipschitz constant of a neural network. 
As the encoder variance $\v{\sigma}^2_{\phi}(\v{x})$ increases, the variance of the decoder function $\mathrm{Var}_{q_\phi(\v{z}|\v{x})}\left( g_{\theta}(\v{z}) \right)$ decreases, which by Proposition~\ref{prop:lipschitz_hermite} will lead to a smaller lower bound on the Lipschitz constant of the decoder.
Due to the computational costs of estimating the Lipschitz constants of networks, we restrict our empirical analysis here to fully-connected VAEs and use \textit{layer-wise} $\mathrm{LipSDP}$  \citep{fazlyab2019} to obtain estimates.
In Appendix ~\ref{app:robustness} we confirm that regulating the encoder variance allows for us to impose a soft constraint of the Lipschitzness of the VAE decoder.
We also show that adding Gaussian noise on data, gives us finer control on the Lipschitz constant of the encoder.
Finally, we also demonstrate that this modulation of the encoder and decoder Lipschitzness is a purveyor of adversarial robustness for VAEs. 

\subsection{Potential Limitations}
\label{sec:limitations}
Our results on Lipschitzness would be stronger if  Proposition~\ref{prop:lipschitz_hermite} offered an upper bound on the Lipschitz constant, in that we could certify a maximum Lipschitz constant. 
Also, the encoder variance does not \textit{directly} affect the frequency content of the encoder, only secondarily.
Noise injections on data are needed to provide full control over the frequency content and Lipschitzness of the VAE networks, whereas
a single parameter to control would have been a stronger result. 
We also recognise that the results in Figure~\ref{fig:robustness_sigma_corr} would have been stronger if the variance of the likelihood degradation results was such that the shading of the curves did not overlap for the different values of $\sigma$ and $\sigma_\phi$.

\section{Conclusion}
Using the lens of Gaussian spaces, we demonstrated that the variance of the latent encodings features as an important parameter for regularising VAE decoders, controlling their Lipschitz constants, and improving their adversarial robustness. By applying the same framework to the VAE encoder we show that simply adding Gaussian noise to VAE inputs offers the same control over the VAE encoder. 
This work lays the foundation for analysing the effect of Gaussian priors on VAEs and offers a novel framework from which to understand the functions learned by a VAE's networks. 
It also paves the way for the study of the effect of \textit{non-Gaussian} priors on VAE networks. 

\section{Acknowledgements}
We thank Professor Svante Janson, Professor Wilfreda Urbina-Romero, and Professor Tom Alberts for their invaluable guidance on Gaussian Spaces.
This research was directly funded by the Alan Turing Institute under Engineering and Physical Sciences Research Council (EPSRC) grant EP/N510129/1.
Alexander Camuto was supported by an EPSRC Studentship.

\bibliographystyle{plainnat}
\bibliography{references.bib}

\clearpage
\newpage
\newpage
\onecolumn
\appendix

\renewcommand\thefigure{\thesection.\arabic{figure}}
\setcounter{table}{0}
\renewcommand{\thetable}{\thesection.\arabic{table}}

{\centering
  {\Large\bfseries Appendix for Variational Autoencoders: a Harmonic Perspective \par}}

\section{Technical Proofs}
\label{app:proofs}

Before we begin the proofs we give an alternative definition of the Hermite polynomials which uses the $k^{\mathrm{th}}$ divergence operator ($\delta^{k}$, \citep{janson_1997}).
This definition greatly simplifies some of our later proofs. 
\begin{equation}\label{eq:hermite_div}
   H_{k} = \delta^{k}1(x), 
\end{equation}
where $1$ is shorthand for the function that is identically 1. 
The divergence operator is defined using $\mathrm{Dom}(\delta^{k})$, which is the subset of functions $g \in L_2(\mathbb{R}, \gamma)$  for which there exists a $c>0$ such that for all functions for which the derivative up to degree $k$ obeys $f^{(k)} \in L_2(\mathbb{R}, \gamma)$: 
\[
\left\lvert \int_\mathbb{R} f^{(k)}(x)g(x)d\gamma(x)\right\rvert \leq c \sqrt{\left\lvert \int_\mathbb{R} f^{(k)}(x)d\gamma(x) \right\rvert}.
\]
$1(x)$ for example, as defined in Equation~\eqref{eq:hermite_div} is in $\mathrm{Dom}(\delta^{k})$. 
The $k^{\mathrm{th}}$ divergence can then be defined as follows. 
If $g \in \mathrm{Dom}(\delta^{k})$ then $\delta^{k} g$ is the unique element of  $L_2(\mathbb{R}, \gamma)$ such that for all functions for which the derivative up to degree $\v{\alpha}$ obeys $f^{(k)} \in L_2(\mathbb{R}, \gamma)$ we have: 
\begin{equation}
    \int_{\mathbb{R}} f(x)\delta^{k}g(x)d\gamma(x) =  \int_{\mathbb{R}} f^{(k)}(x)d\gamma(x).
\end{equation}
For the general Gaussian space, this becomes by substitution of variables:
\begin{equation}
   H_{k} = \delta^{k}1(\hat{x}),\qquad \hat{x} = (x - \mu)/\sigma.
\end{equation}

In the multivariate case, for the general Gaussian space with diagonal standard deviation, we have that $\mathcal{H}_{\v{\alpha}} = \delta^{\v{\alpha}}1(\v{\hat{x}}), \ \v{\hat{x}} = (\v{x}-\v{\mu})\v{S}^{-1}$,
where $\delta$ now is the vector-valued divergence operator indexed by the multi-index $\v{\alpha}$.

\subsection{Proof of Proposition~\ref{prop:general_gauss_space}}
\label{app:general_gauss_space}
\begin{proof}
Due to the scaling and centering relation for Gaussians a simple change of variables $\hat{x}=(x-\mu)/\sigma$ in Equation~\eqref{eq:ortho_hermite} gives us the first part of the proof. 
Thus 
\[
\expect_{\gamma(\hat{x})}[H_k(\hat{x})H_m(\hat{x})] = k! \mathbf{1}\{m=k\},
\]
and thus $\left\{\frac{1}{k!}H_k: k \geq 0 \right\}$ is orthonormal in $L_2(\mathbb{R}, \mathcal{N}(\mu,\sigma^2))$ because $\gamma(\hat{x}) = \mathcal{N}(\mu,\sigma^2)$. 

We know that for $k \geq 0$ the polynomial $H_k$ has degree $k$, hence it suffices to demonstrate that the mononomials of degree $k$, $\{y^k: k\in\mathbb{N}, \hat{x}=(x-\mu)/\sigma \}$ are dense in  $L_2(\mathbb{R}, \gamma(\hat{x}))$ to show that  $\left\{\frac{1}{\v{\alpha}!}H_k: k \geq 0 \right\}$ is a basis for $L_2(\mathbb{R}, \gamma(\hat{x}))$.

The Elementary Hahn-Banach theorem  shows that if $g \in L_2(\mathbb{R}, \gamma)$ is such that $\int_{\mathbb{R}} g(\hat{x}) \hat{x}^{k} d \gamma(\hat{x})=0, \forall k \in \mathbb{N}^+$ then it suffices to show that $g=0$ almost everywhere to demonstrate that the mononomials $\hat{x}^{k}$ are a dense subspace of  $L_2(\mathbb{R}, \gamma(\hat{x}))$.
This proof generally follows that used for the standard Gaussian measure. 
Assume we have a $g$ that satisfies $\int_{\mathbb{R}} g(\hat{x}) \hat{x}^k d \gamma(\hat{x})=0$.
By the series expansion of the exponential function we know that for all $t \in \mathbb{R}$
\begin{align*}
 \int_\mathbb{R} g(\hat{x})e^{ity}d\gamma(\hat{x}) =  \lim_{m \to \infty} \sum_{k=0}^m\frac{(it)^k}{k!}  \int_\mathbb{R} g(\hat{x})\hat{x}^kd\gamma(\hat{x}) = 0 .
\end{align*}

Thus we have that $\int_\mathbb{R} g(\hat{x})e^{ity}d\gamma(\hat{x})=0, \forall t$.
Because $e^{ity}$ is injective we have that  $g(\hat{x}) = 0, \forall \hat{x} \in \mathbb{R}$.

\end{proof}

\subsection{Proof of Proposition~\ref{prop:hermite_coeffs}} \label{app:hermite_coeffs}

\begin{proof}

We begin with the univariate case, from which we build up to the multivariate case. 
Here we use the divergence operator definition of the Hermite polynomials, seen in Equation~\eqref{eq:hermite_div}. 
In our case by substitution of variables $H_{k}(\hat{x}) = \delta^{k}1(\hat{x})$ and we can express $x$ as $x = y\sigma + \mu$, thus: 
 \[
    \int_{\mathbb{R}} f(y\sigma + \mu)H_{k}(\hat{x})d\gamma(\hat{x}) =  \int_{\mathbb{R}} f(y\sigma + \mu)\delta^{k}1(\hat{x})d\gamma(\hat{x}) =  \int_{\mathbb{R}} \sigma^k f^k(\hat{x}\sigma + \mu)d\gamma(\hat{x}).
\]
As mentioned in the main paper, in the multivariate case, the divergence operator is an almost exact analogue to that of the univariate case. 
Due to the tensorisation of bases to form $\mathcal{H}$ the indices $\alpha$ simply become a multi-index and we recover  that $\hat{f}(\v{\alpha}) =  \expect_{\gamma(\v{\hat{x}})}\left[\v{S}^{\v{\alpha}} f^{(\v{\alpha})}(\v{x})\right]$.

\end{proof}
\subsection{Proof of Theorem~\ref{thm:var_fourier}}\label{app:var_fourier}

\begin{proof}

By Proposition~\ref{prop:hermite_coeffs} we know that we can express a function $f \in L_2(\mathbb{R}^n, \mathcal{N}(\v{\mu}, \v{S}^2))$ that is infinitely differentiable as: 
\[f = \sum_{\v{\alpha} \in \mathbb{N}^n} \frac{1}{\v{\alpha}!} (\mathrm{diag}(\v{S}))^{\v{\alpha}}  \expect_{\gamma(\v{\hat{x}})}\left[ f^{(\v{\alpha})}(\v{x})\right]\mathcal{H}_{\v{\alpha}}, \qquad  \gamma(\v{\hat{x}}) = \mathcal{N}(\v{x}|\v{\mu}, \v{S}^2).
\]

Now note that the expectation of $f$ is simply the first Hermite coefficient with degree 0: 
\[
\expect_{\gamma(\v{\hat{x}})}\left[f\right] = \sum_{|\v{\alpha}|=0 } \frac{1}{\v{\alpha}!} (\mathrm{diag}(\v{S}))^{\v{\alpha}}  \expect_{\gamma(\v{\hat{x}})}\left[ f^{(\v{\alpha})}(\v{x})\right]\mathcal{H}_{\v{\alpha}}, \qquad  \gamma(\v{\hat{x}}) = \mathcal{N}(\v{x}|\v{\mu}, \v{S}^2).
\]

Now taking the variance of $f$: 
\begin{align*}
   \mathrm{Var(f)} &= \expect_{\gamma(\v{\hat{x}})}\left[\left(f -\expect_{\gamma(\v{\hat{x}})}\left[f\right] \right)^2\right] \\
   &= \expect_{\gamma(\v{\hat{x}})}\left[\left( \sum_{\v{\alpha} \in \mathbb{N}^n} \frac{1}{\v{\alpha}!} (\mathrm{diag}(\v{S}))^{\v{\alpha}}  \expect_{\gamma(\v{\hat{x}})}\left[ f^{(\v{\alpha})}(\v{x})\right]\mathcal{H}_{\v{\alpha}}  -\sum_{|\v{\alpha}|=0 } \frac{1}{\v{\alpha}!} (\mathrm{diag}(\v{S}))^{\v{\alpha}}  \expect_{\gamma(\v{\hat{x}})}\left[ f^{(\v{\alpha})}(\v{x})\right]\mathcal{H}_{\v{\alpha}} \right)^2\right] \\
   &= \expect_{\gamma(\v{\hat{x}})}\left[\left( \sum_{|\v{\alpha}| \geq 1} \frac{1}{\v{\alpha}!} (\mathrm{diag}(\v{S}))^{\v{\alpha}}  \expect_{\gamma(\v{\hat{x}})}\left[ f^{(\v{\alpha})}(\v{x})\right]\mathcal{H}_{\v{\alpha}}  \right)^2\right].
\end{align*}

Recall that the bases $\mathcal{H}_{\v{\alpha}}$ are orthogonal such that 
\begin{equation*}
    \expect_{\gamma(\v{\hat{x}})}[\mathcal{H}_{\v{\alpha}}(\v{\hat{x}})\mathcal{H}_m(\v{\hat{x}})] = \v{\alpha}!\{m=\v{\alpha}\}.
\end{equation*}
As such we have that: 
\begin{align*}
   \mathrm{Var(f)} &= \expect_{\gamma(\v{\hat{x}})}\left[\left( \sum_{|\v{\alpha}| \geq 1} \frac{1}{\v{\alpha}!} (\mathrm{diag}(\v{S}))^{\v{\alpha}}  \expect_{\gamma(\v{\hat{x}})}\left[ f^{(\v{\alpha})}(\v{x})\right]\mathcal{H}_{\v{\alpha}}  \right)^2\right] \\
   &= \expect_{\gamma(\v{\hat{x}})}\left[ \sum_{|\v{\alpha}| \geq 1} \left(\frac{1}{\v{\alpha}!} (\mathrm{diag}(\v{S}))^{\v{\alpha}}  \expect_{\gamma(\v{\hat{x}})}\left[ f^{(\v{\alpha})}(\v{x})\right]\mathcal{H}_{\v{\alpha}}\right)^2  \right] \\
   &= \sum_{|\v{\alpha}| \geq 1 } \frac{(\mathrm{diag}(\v{S}))^{2\v{\alpha}}}{\v{\alpha}!} \left|\expect_{\gamma(\v{\hat{x}})}\left[ f^{(\v{\alpha})}(\v{x})\right]\right|^2.
\end{align*}
This completes the first part of the proof. 

For the second part we focus on terms of the form 
\[
\expect_{\gamma(\v{\hat{x}})}\left[ f^{(\v{\alpha})}(\v{x})\right] = \int_{\mathbb{R}^n} f^{(\v{\alpha})}(\v{\hat{x}}\v{S} + \v{\mu})d\gamma(\v{\hat{x}}) = \int_{\mathbb{R}^n} f^{(\v{\alpha})}(\v{\hat{x}}\v{S} + \v{\mu})\gamma(\v{\hat{x}}) d\v{\hat{x}}.
\]
Both $f$ and $\gamma$ are $L_2$ integrable with respect to the Lebesgue measure. 
As such we can directly apply Plancherel's theorem to demonstrate that 
\begin{align*}
\expect_{\gamma(\v{\hat{x}})}\left[ f^{(\v{\alpha})}(\v{x})\right]  &= \int_{\mathbb{R}^n} f^{(\v{\alpha})}(\v{\hat{x}}\v{S} + \v{\mu})\gamma(\v{\hat{x}}) d\v{\hat{x}} \\ 
&= \int_{\mathbb{R}^n} f^{(\v{\alpha})}(\v{x})\gamma((\v{x} - \v{\mu})\v{S}^{-1}) d\v{x} \\ 
&= 
    \int_{\mathbb{R}^n}
   (i\v{\omega})^{\v{\alpha}}\mathcal{F}(\v{\omega})\overline{\mathrm{det}(\v{S})\mathcal{G}(\v{\omega}\v{S})e^{-i\v{\omega}\v{\mu}\v{S}^{-1}}}d\v{\omega} \\
&= 
    \int_{\mathbb{R}^n}
   (i\v{\omega})^{\v{\alpha}}\mathcal{F}(\v{\omega})\overline{\mathcal{P}(\v{\omega})}d\v{\omega}.
\end{align*}
where $\mathcal{P}$ is the Fourier transform of the Gaussian measure $\mathcal{N}(\mu,\sigma^2)$ given by 
$\overline{\mathcal{P}(\v{\omega})} = \mathrm{det}(\v{S})\mathcal{G}(\v{\omega})e^{-i\v{\omega}\v{\mu}\v{S}^{-1}}$ (where $\mathcal{G}$ is the Fourier transform of the standard Gaussian measure $\gamma$) and $\mathcal{F}$ is the Fourier transform of $f$. 
This stems from the fact that the Fourier transform of the $\v{\alpha}^\mathrm{th}$ derivative of a function is simply $(i\v{\omega})^{\v{\alpha}} \mathcal{F}(\v{\omega})$

From this we obtain that 
\begin{align*}
\mathrm{Var(f)} &= \sum_{|\v{\alpha}| \geq 1 } \frac{(\mathrm{diag}(\v{S}))^{2\v{\alpha}}}{\v{\alpha}!} 
    \left|\int_{\mathbb{R}^n} (i\v{\omega})^{\v{\alpha}}\mathcal{F}(\v{\omega})\overline{\mathcal{P}(\v{\omega})}d\v{\omega}\right|^2 \\
    &=  \sum_{|\v{\alpha}| \geq 1 } \frac{(\mathrm{diag}(\v{S}))^{2\v{\alpha}}}{\v{\alpha}!} 
    \left|\int_{\mathbb{R}^n} (\v{\omega})^{\v{\alpha}}\mathcal{F}(\v{\omega})\overline{\mathcal{P}(\v{\omega})}d\v{\omega}\right|^2, \v{\alpha} \in \mathbb{N}^n.
\end{align*}

\end{proof}
\subsection{Proof of Proposition~\ref{prop:lipschitz_hermite}}\label{app:lipschitz_hermite}
\begin{proof}
We begin with a proof for a univariate Gaussian space. 
We once again use the definition of the Hermite polynomials in terms of the divergence operator. 
Note that $H_\alpha = \delta H_{\alpha-1}$. 
The variance of a function $f \in L_2(\mathbb{R}, \mathcal{N}(\mu, \sigma^2))$ can be expressed as the sum of Hermite coefficients squared: 
\begin{equation}
    \mathrm{Var}(f) = \sum_{\alpha =1}^{\infty}\frac{1}{\alpha!} |\hat{f}(\alpha)|^2= \sum_{\alpha =1}^{\infty} \frac{1}{\alpha!}\left|\expect_{\gamma(\hat{x})}\left[f(\sigma \hat{x} + \mu )H_\alpha(\hat{x})\right]\right|^2.
\end{equation}
Using the definition of the divergence operator we have that: 
\begin{align}
    \mathrm{Var}(f) &=  \sum_{\alpha =1}^{\infty} \frac{1}{\alpha!}\left|\expect_{\gamma(\hat{x})}\left[\sigma f^{(1)}(\sigma \hat{x} + \mu )H_{\alpha-1}(\hat{x})\right]\right|^2 \\ &\leq \sum_{\alpha =1}^{\infty} \frac{1}{(\alpha -1)!}\left|\expect_{\gamma(\hat{x})}\left[\sigma f^{(1)}(\sigma \hat{x} + \mu )H_{\alpha-1}(\hat{x})\right]\right|^2
    \\ &= \sum_{\alpha =0}^{\infty} \frac{\sigma^2}{\alpha!}\left|\expect_{\gamma(\hat{x})}\left[ f^{(1)}(\sigma \hat{x} + \mu )H_{\alpha}(\hat{x})\right]\right|^2.
\end{align}
As $f^{(1)}$ is a member of the Gaussian space, the last equation is simply the dot product of $f^{(1)}$ with itself, weighted by $\sigma^2$.
Thus: 
\begin{align}
    \mathrm{Var}(f) &= \sigma^2 \langle f^{(1)}, f^{(1)} \rangle.
\end{align}
By definition of the Lipschitz constant: 
\[
L \leq \sup_{x,x' \in \mathbb{R^n}, x \neq x'}\frac{|f(x) - f(x')|}{\|x-x'\|}.
\]
We then have 
\begin{align}
    \mathrm{Var}(f) &\leq \sigma^2 L^2.
\end{align}

For the multivariate case, we work in reverse. 
We know that $L^2 \geq \|D f\|^2$ where $D f$ is the Jacbobian of the function with respect to the input such that  

\[\left[\|D f\|^2\right] = \sum_{|\v{\alpha}|=1} <f^{(\v{\alpha})}, f^{(\v{\alpha})}>
\]
As by assumption each $f^{(\v{\alpha})}, |\v{\alpha}|=1$ is a member of the Gaussian space we have that 
\begin{align}
    \|D f\|^2 &= \sum_{|\v{\alpha}|=1} \sum_{\beta \in \mathbb{N}^n} \frac{1}{\beta!} \left|\expect_{\gamma(\v{\hat{x}})} \left[f^{(\v{\alpha})}(\v{S}\v{\hat{x}} + \v{\mu})\mathcal{H}_{\beta}(\v{\hat{x}}) \right]\right|^2 \\
    &= \sum_{|\v{\alpha}|=1} \sum_{\beta \in \mathbb{N}^n} \frac{1}{\beta!} \left|\expect_{\gamma(\v{\hat{x}})} \left[\v{S}^{-1}f(\v{S}\v{\hat{x}} + \v{\mu})\delta^{\v{\alpha}}\mathcal{H}_{\beta}(\v{\hat{x}}) \right]\right|^2 \\
    &= \sum_{|\v{\alpha}|=1} \sum_{\beta \in \mathbb{N}^n} \frac{1}{\beta!} \left|\expect_{\gamma(\v{\hat{x}})} \left[\v{S}^{-1}f(\v{S}\v{\hat{x}} + \v{\mu})\mathcal{H}_{\beta + \v{\alpha}}(\v{\hat{x}}) \right]\right|^2 \\
    &\geq \sum_{|\beta| \geq 1} \frac{1}{\beta!} \left|\expect_{\gamma(\v{\hat{x}})} \left[\v{S}^{-1}f(\v{S}\v{\hat{x}} + \v{\mu})\mathcal{H}_{\beta}(\v{\hat{x}}) \right]\right|^2, \beta \in \mathbb{N}^n.
\end{align}
This last term is simply $\mathrm{Var}(f)/\|\v{S}\|^{2}$ (see the proof of Theorem~\ref{thm:var_fourier} for why this is). 
Thus we have that $L^2\|\v{S}\|^{2} \geq \mathrm{Var}(f)$. 
This concludes the proof.

\end{proof}

\section{Multivariate Hermite Coefficients for a Gaussian measure with full Covariance }\label{app:hermite_full_cov}

Let our Gaussian space be equipped with Gaussian measure $\mathcal{N}(\v{\mu}, \v{C})$, where $\v{C}$ is a full covariance matrix. 
By definition $\v{C}$ is symmetric and positive definite, thus there exists an orthogonal matrix $\v{O}$ such that: 
\[
\v{O}\v{C}\v{O}^T =\v{D}, 
\]
where $\v{D}$ is some diagonal matrix. 
Let $\v{x} \sim \mathcal{N}(\v{\mu}, \v{C})$, we know that:
\[\v{\hat{x}} = (\v{x}-\v{\mu})\v{O}^T\v{D}^{-\frac{1}{2}} \sim \mathcal{N}(\v{0}, \v{I}).\]
By substitution of variables for the Hermite polynomials in the case of a diagonal covariance, we have that the Hermite polynomials for the measure $\mathcal{N}(\v{\mu}, \v{C})$ are given by:
\[\mathcal{H}_{\v{\alpha}}((\v{x}-\v{\mu})\v{O}^T\v{D}^{-\frac{1}{2}})=\mathcal{H}_{\v{\alpha}}(\v{\hat{x}})=\prod_i H_{\alpha_i}(\hat{x}_i).\]

\section{Implementation Details}\label{app:architecture}

The architecture of VAEs with dense layers is presented in text. 
For convolutional networks the encoder layers had the following number of filters in order: 
$\{64,64,128,128,512\}$. 

The mean and variance of the amortised posteriors are the output of dense layers acting on the output of the purely convolutional network, where the number of neurons in these layers is equal to the dimensionality of the latent space $\mathcal{Z}$.

Similarly, for the decoders ($p(\v{x}|\v{z})$) of all our models we also used purely convolutional networks with 6 deconvolutional layers. 
The layers had the following number of filters in order: 
$\{512,128,128,64,64,3\}$. 
The mean of the likelihood $p(\v{x}|\cdot)$ was directly encoded by the final de-convolutional layer.
The variance of the decoder was fixed to 0.1. 

To train models we used ADAM \citep{Kingma2015} with default parameters, a learning rate of 0.001, and a batch size of 1024. 
All data was preprocessed to fall on the interval -1 to 1.

All models were trained with an \textit{Azure Cloud Standard NC6} machine with a single  \textit{NVIDIA Tesla K80}. 
\newpage
\section{Additional Results}

\subsection{Fourier Transform of Gaussian Measure}
\begin{figure}[h!]
\centering
\includegraphics[width=0.32\textwidth]{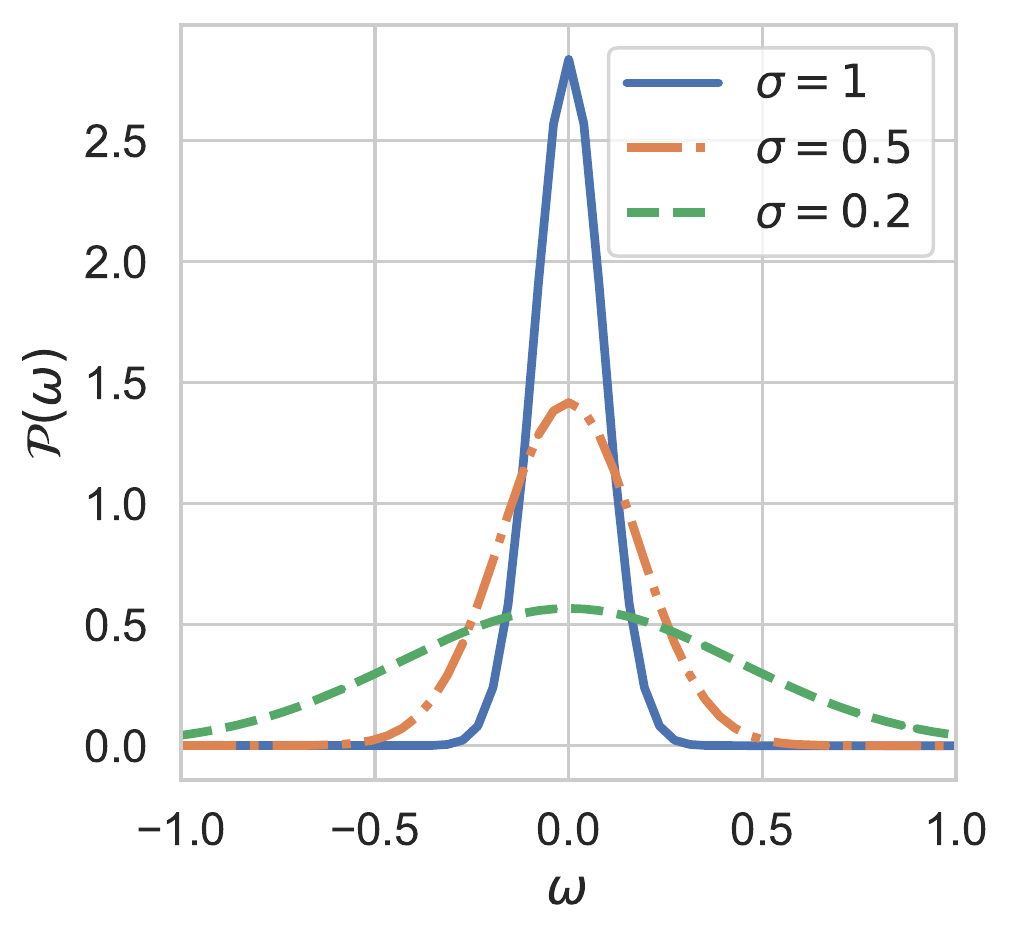}
\hspace{1cm}
\includegraphics[width=0.32\textwidth]{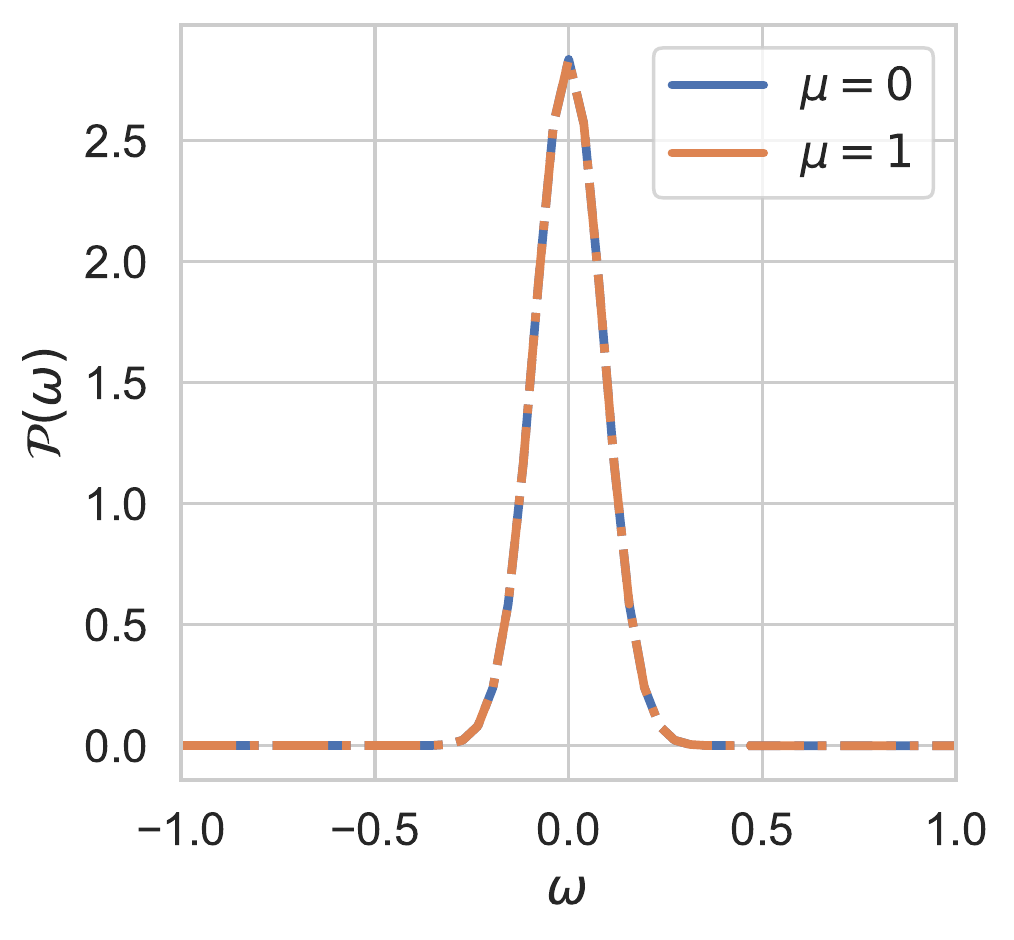}
\caption{\label{fig:gaussian_ft}[left] Fourier transform of the Gaussian measure $\mathcal{N}(0,\sigma^2)$ for varying $\sigma$. 
[right] Fourier transform of the Gaussian measure $\mathcal{N}(\mu,1)$ for varying $\mu$. 
Clearly as $\sigma$ decreases the spectrum gains high frequencies, whereas altering $\mu$ has no effect. }
\end{figure}

\subsection{Fourier Spectrum of Multivariate VAEs}

\begin{figure*}[h!]
\centering
    \includegraphics[width=0.3\textwidth]{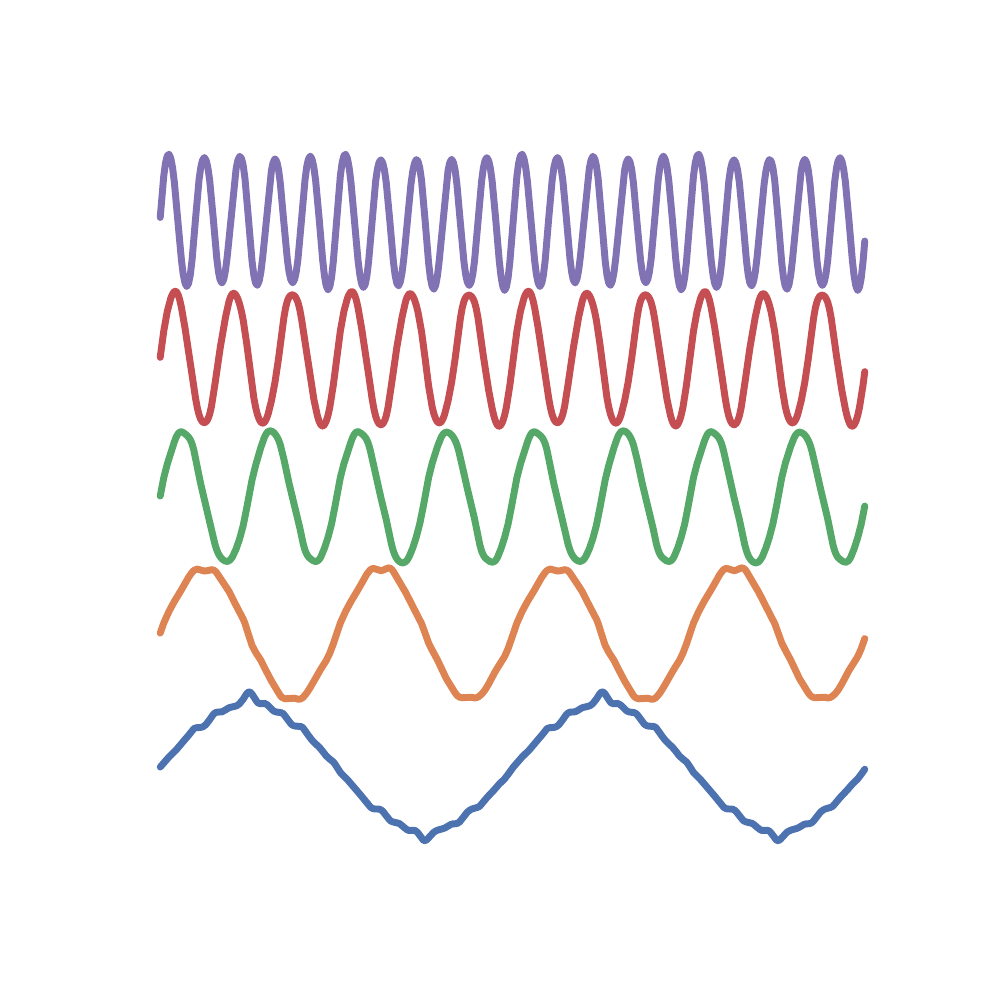}
    \includegraphics[width=0.3\textwidth]{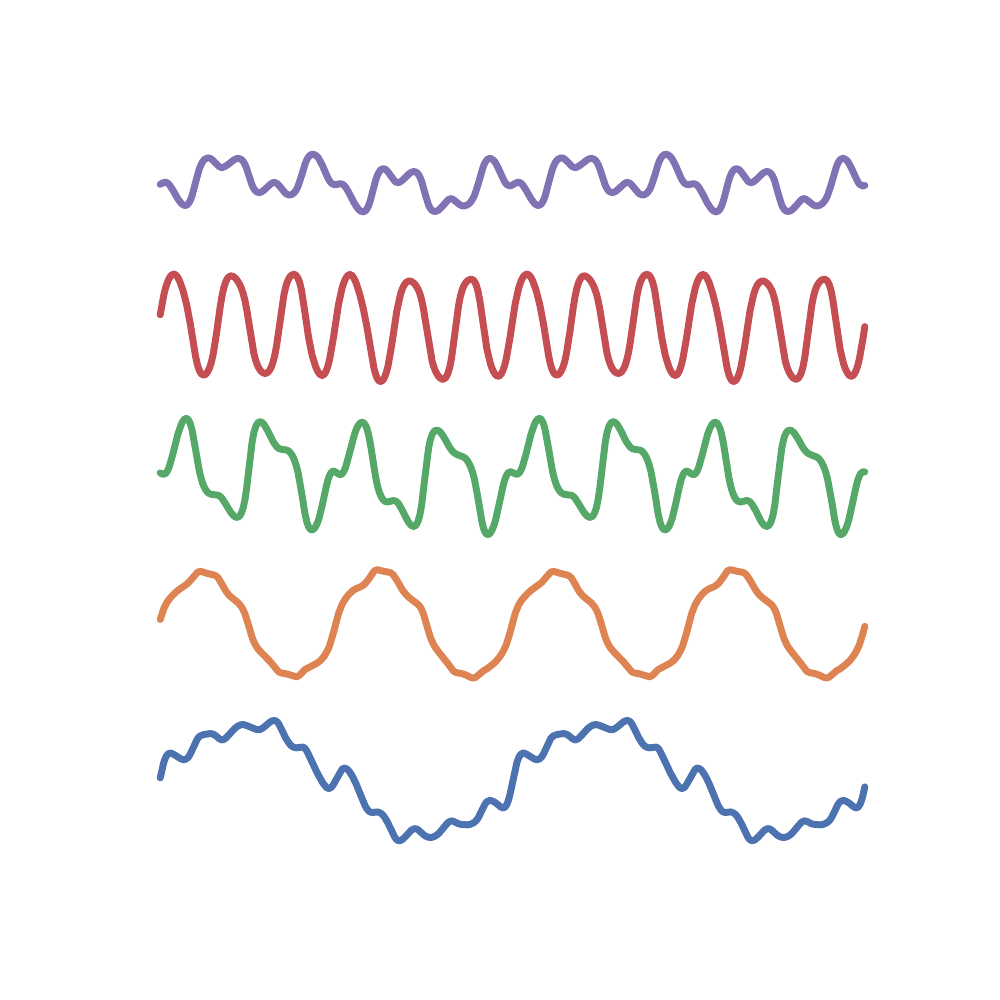}
    \includegraphics[width=0.3\textwidth]{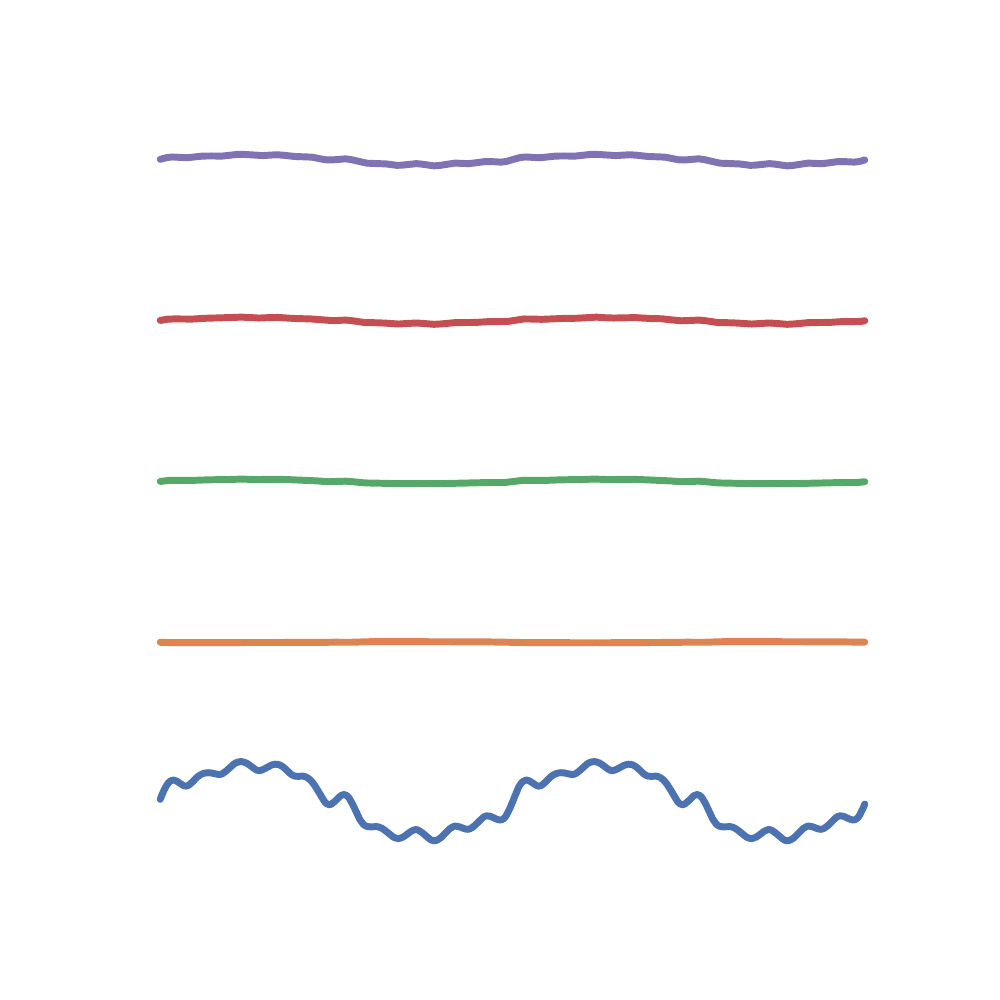} \\ 
    \includegraphics[width=0.3\textwidth]{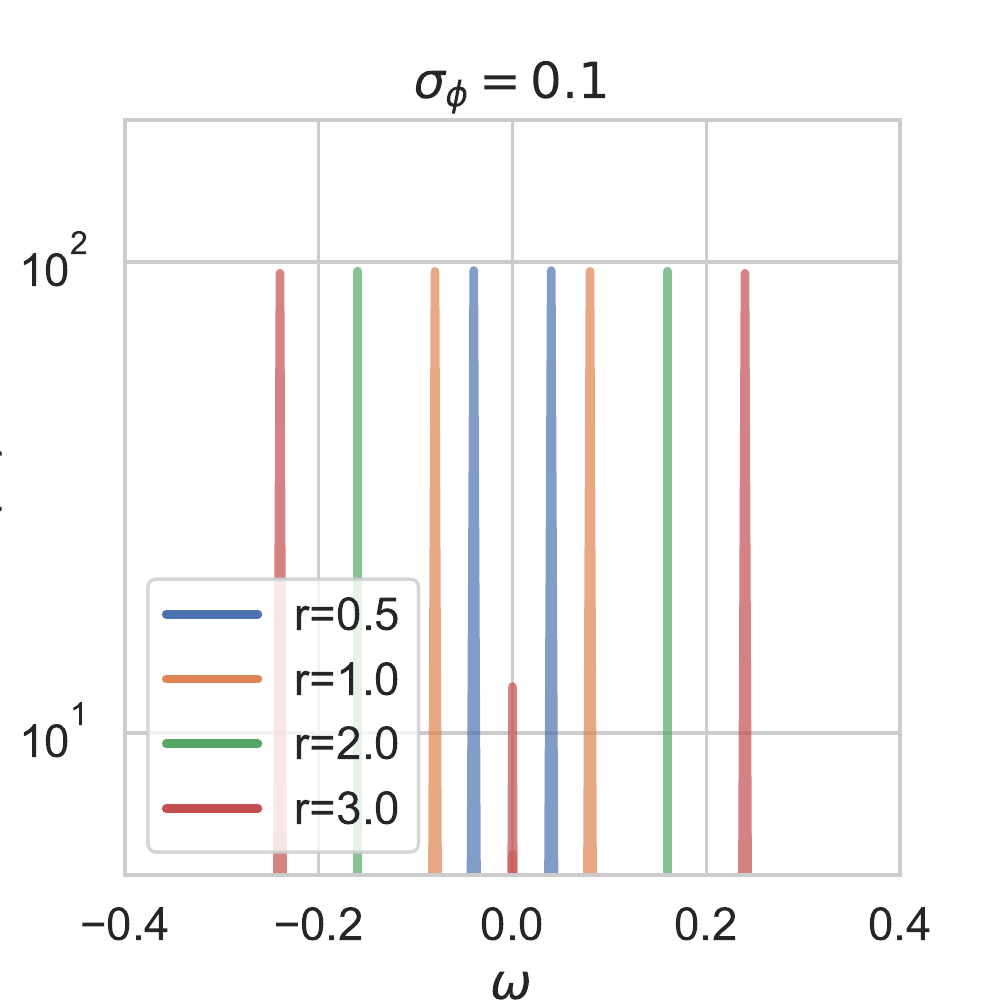}
    \includegraphics[width=0.3\textwidth]{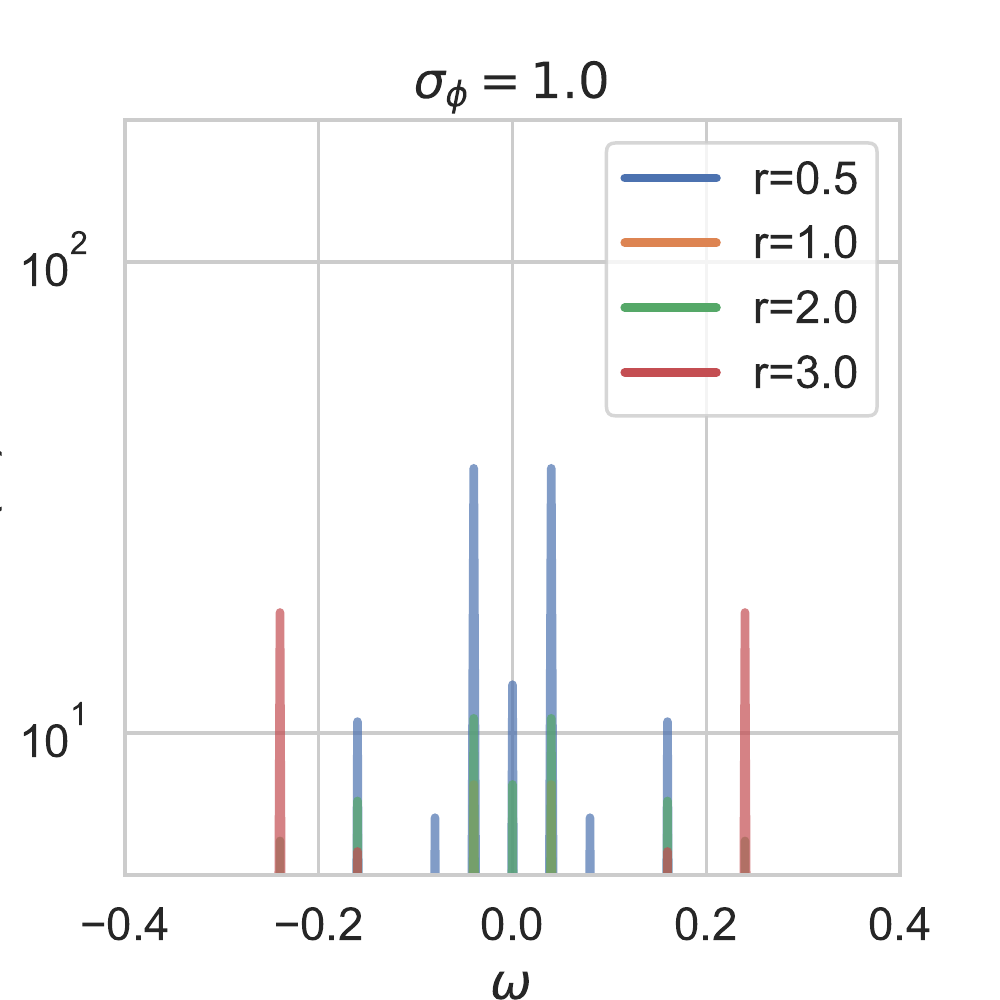}
    \includegraphics[width=0.3\textwidth]{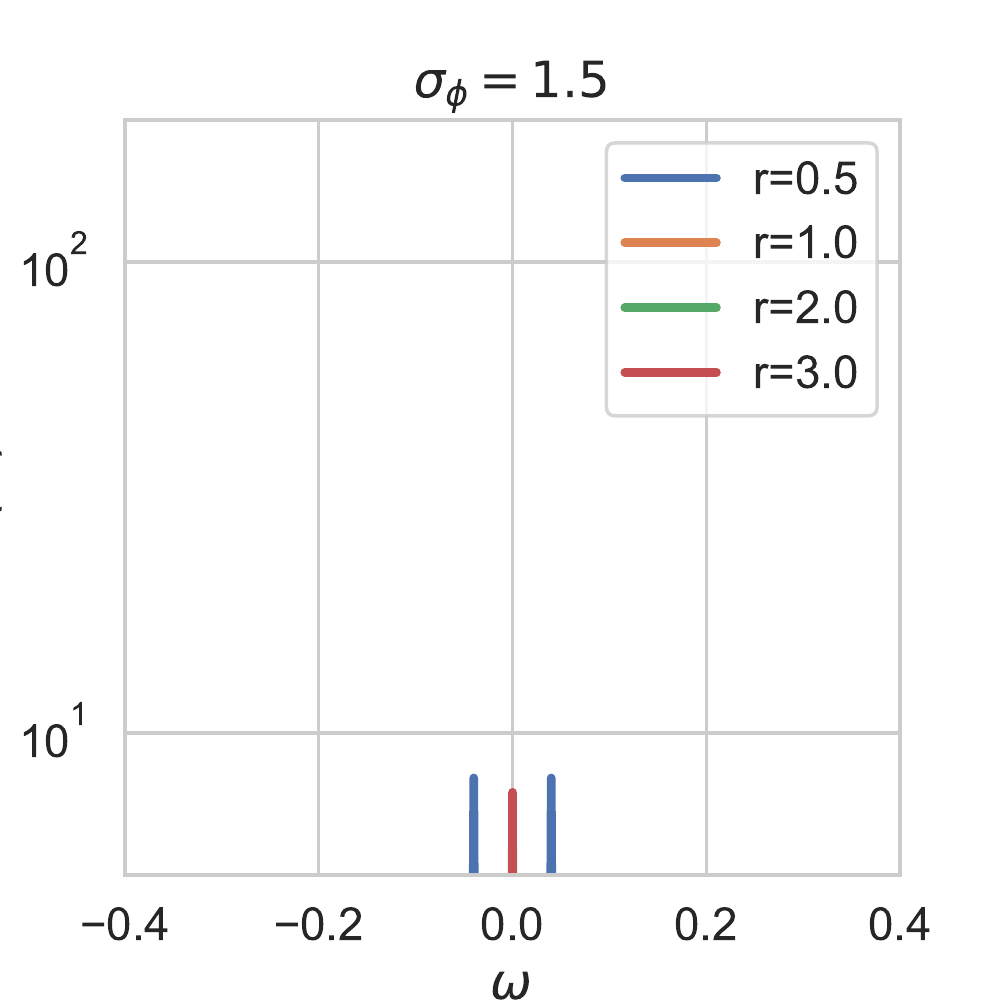}
    \caption{
    We train VAEs with 3-dense layers (each with 256 units), with $\mathrm{ReLU}$ activation functions in their networks, on multivariate data consisting of 5 sinusoids $y=\mathrm{sin}(2\pi rt)$, $r \in \{0.5,1.0,2.0,3.0, 5.0\}$, $t\in [-1,1]$, with a 1 dimension latent space $\mathcal{Z}$. As before we fix the encoder $\sigma_\phi$. 
    \textbf{[top row]} We show plots of the regressed sinusoids for different values of $\sigma_\phi$. \textbf{[bottom row]} Below we show an FFT for each regressed sinusoid. As $\sigma_\phi$ increases we clearly lose some of the mid to high level frequencies.
    } 
    \label{fig:frequency_illus_multi}
\end{figure*}

\subsection{Gaussian Spaces and VAE Lipschitzness} 
\label{app:robustness}
We know that by Proposition~\ref{prop:lipschitz_hermite} the variance of a function (that is once differentiable) provides a relatively tight lower bound on the Lipschitz constant of a neural network. 
This lower bound also increases as the variance of the Gaussian measure decreases. 
In the Supplement we demonstrate that this modulation of the encoder and decoder Lipschitzness is a purveyor of adversarial robustness for the VAE networks. 

In previous sections we demonstrated that larger encoder variances induce smaller variance decoder functions, with lower frequency components.
As the encoder variance $\v{\sigma}^2_{\phi}(\v{x})$ increases, the variance of the decoder function $\mathrm{Var}_{q_\phi(\v{z}|\v{x})}\left( g_{\theta}(\v{z}) \right)$ decreases, which by Proposition~\ref{prop:lipschitz_hermite} will lead to a smaller lower bound on the Lipschitz constant of the decoder.
Due to the computational costs of estimating the Lipschitz constants of networks, we restrict our empirical analysis here to fully-connected VAEs and use \textit{layer-wise} $\mathrm{LipSDP}$  \citep{fazlyab2019} to obtain estimates.
In Table~\ref{table:lip_illus_decoder} we confirm that regulating the encoder variance allows for us to impose a soft constraint of the Lipschitzness of the VAE decoder.
We also show that adding Gaussian noise on data, as motivated in the previous sections by its effect on $\mathrm{Var}(\mu_\phi)$, gives us finer control on the Lipschitz constant of the encoder.

\begin{table}[h]
  \caption{\small The Lipschitz constants ($L$) of VAEs' \textbf{[left]} decoder networks ($g_\theta$) when trained with fixed encoder scale ($\sigma_\phi$) and \textbf{[right]}  encoder networks ($\mu_\phi$) when trained with $\sigma$-scale Gaussian noise injection on inputs and fixed encoder variance $\sigma_\phi$=$0.5$. We train fully-connected VAEs  with the $\mathrm{Sigmoid}$ activation in their networks; on $\mathrm{sinc}(5t)$ ($\mathrm{dim}(\mathcal{Z})=1$), on CelebA ($\mathrm{dim}(\mathcal{Z})=64$), and CIFAR10 ($\mathrm{dim}(\mathcal{Z})=64$). 
  Each network has 3 dense layers each with 256 units. 
   As $\sigma_\phi$ decreases the Lipschitz constant ($L$) increases, as predicted by Proposition~\ref{prop:lipschitz_hermite}. As $\sigma$ decreases, the Lipschitz constant ($L$) of the encoder mean ($\mu_\phi$) increases. ($\pm$) is the std. dev. over 3 random seeds.}
\label{table:lip_illus_decoder}
  \centering

  \begin{tabular}{c|ccc|ccc}
    \toprule
\multirow{2}{*}{Data}    &\multicolumn{3}{c|}{$L(g_\theta; \sigma_\phi$)}&\multicolumn{3}{c}{$L(\mu_\phi;\sigma)$}\\
                            & $\sigma_{\phi} = 1.0$ & $\sigma_{\phi} = 0.5$ & $\sigma_{\phi} = 0.1$& $\sigma=1.0$ & $\sigma=0.5$ & $\sigma=0.0$\\
        \midrule

                                $ \mathrm{sinc}$ & $2.2 \pm 0.2 $  & $5.2 \pm 0.3 $ & $17.9 \pm 3.2 $ & $13.9 \pm 2.7$  & $24.6 \pm 1.7$ & $29.8 \pm 2.2$ \\
                                $ \mathrm{CelebA}(10^4)$ & $7.5 \pm 1.1$  & $12.0 \pm 0.5$ & $13.7 \pm 1.2$ & $1.4 \pm 0.1$  & $1.6 \pm 0.1$ & $1.8 \pm 0.1$ \\
                                
                   $ \mathrm{CIFAR10}(10^2)$ & $17.9 \pm 1.2$  & $19.1 \pm 1.2$ & $27.3 \pm 0.3$ & $4.7 \pm 0.2$  & $5.6 \pm 0.6$ & $8.5 \pm 0.8$ \\
    
    \bottomrule
  \end{tabular}
\end{table}

\paragraph{Lipschitzness and Robustness}

\begin{figure}[h!]
\centering
\includegraphics[width=0.33\textwidth]{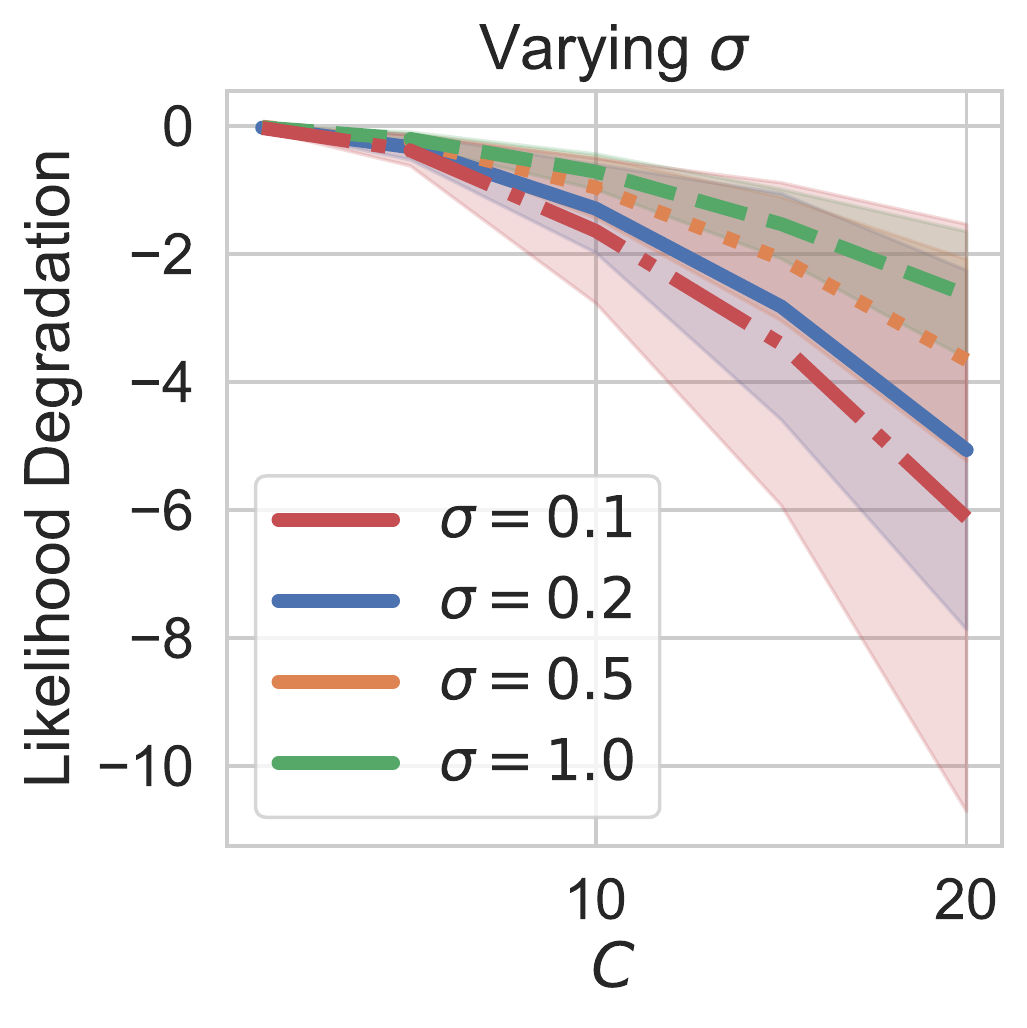}
\hspace{1cm}
\includegraphics[width=0.33\textwidth]{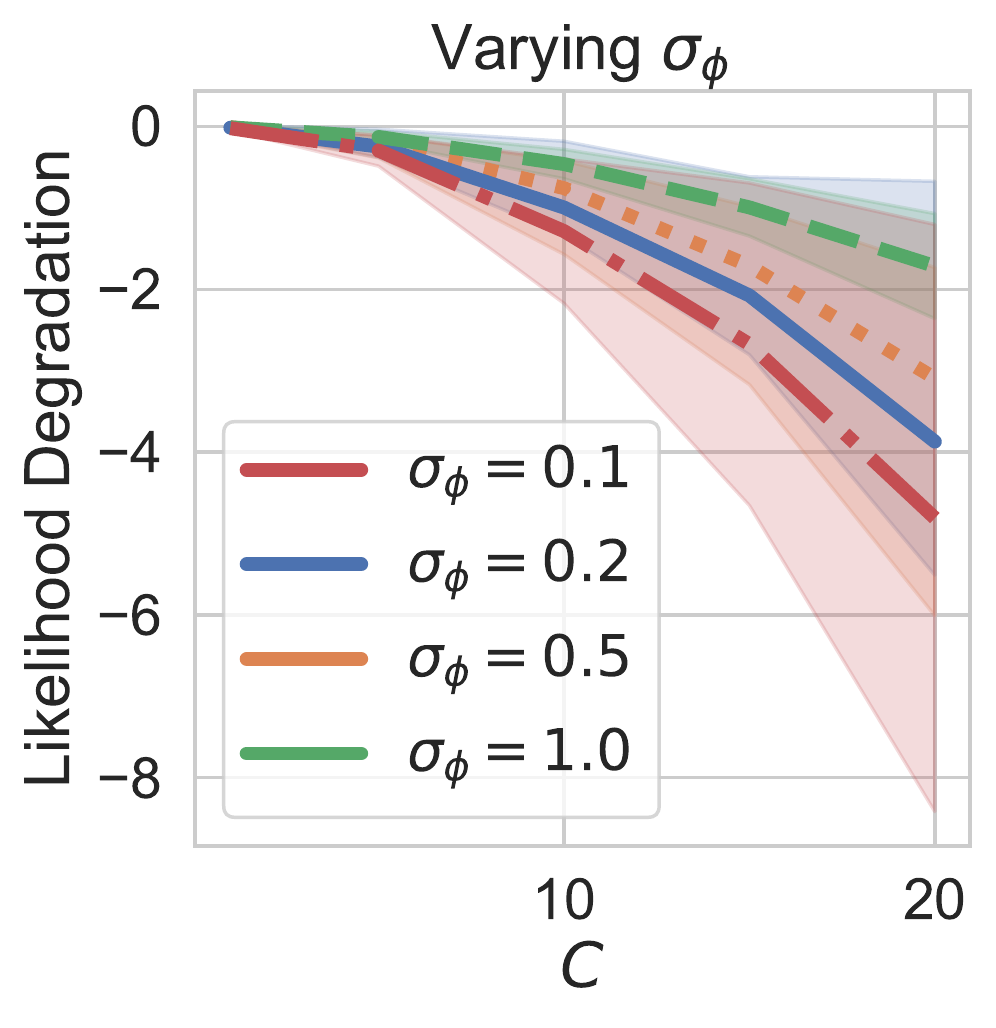}
\caption{Both plots show the relative log likelihood degradation resulting from a `maximum-damage' \citep{Camuto2020_2} adversarial attack against the maximum attack norm $C$. As $\sigma$, the variance of isotropic Gaussian noise added to data, and  $\sigma_\phi$, the fixed posterior variance, increase, the degradation lessens as $C$ increases, highlighting that both these parameters improve robustness. Shading corresponds to the variance of the likelihood degradation over 25 points from $\mathrm{CIFAR10}$.
}
\label{fig:robustness_sigma_corr}
\end{figure}

Recent work shows that larger encoder variances and smaller encoder and decoder network Lipschitz constants are core contributors to the robustness of VAEs to adversarial attack \citep{alex2020theoretical, barrett2021}.  
Whereas this theory has viewed the network Lipschitz constants and the encoder variance as distinct parameters to control to attain robustness; our harmonic analysis shows that they are in fact intrinsically linked.  
This means that the encoder variance can serve as a single parameter on which we can act to improve the robustness of VAEs to adversarial attack, in that it also affects the Lipschitz constant of the decoder.
Further, adding noise to the VAE input data gives us similar control on the Lipschitz constant of the encoder mean $\mu_\phi$.
As such, fixing and modulating both the encoder variance and the variance of the noise on data allows for the imposition of soft Lipschitz constraints on the VAE networks that improve adversarial robustness.

We consider an adversary trying to distort the input data to maximally disrupt a VAE's output, as in \textit{maximum damage} attacks \citep{Camuto2020_2, barrett2021}. 
The adversary maximises, with respect to a perturbation on data $\v{\delta}_x$, constrained in norm by a constant C, the distance between the VAE reconstruction and data $\v{x}$
\begin{equation}\label{eq: max_damage_attack}
 \v{\delta}_x^* =\argmax_{\|\v{\delta}_x\|_2 \leq C}\big(||g_{\theta}(\v{\mu}_\phi(\v{x} + \v{\delta}_x)+ \v{\eta}\sigma_\phi(\v{x} + \v{\delta}_x)) - g_{\theta}(\v{\mu}_\phi(\v{x}))||_2\big).
\end{equation}
Given an embedding $\v{z}^*$ formed from the mean encoding of $\v{x} + \v{\delta}_x$, we measure the likelihood of the original point $\v{x}$ and quantify the degradation in model performance as the relative log likelihood degradation ($|\log p_\theta(\v{x}|\v{z}^*) - \log p_\theta(\v{x}|\v{z})|/\log p_\theta(\v{x}|\v{z})$), where $\v{z}$ is the embedding of $\v{x}$. 

Figure \ref{fig:robustness_sigma_corr} shows that as the variance of noise on data ($\sigma$) and the fixed encoder variance ($\sigma_\phi$) increase, this degradation lessens for the VAEs trained on the CIFAR10 dataset in Table \ref{table:lip_illus_decoder}, indicating less damaging attacks.
In tandem, Table \ref{table:lip_illus_decoder} also shows that the increase of both these parameters decreases the Lipschitz constants of the encoder and decoder network for these VAEs.
Thus we confirm the analysis above (and the recent theory on the adversarial robustness of VAEs \citep{alex2020theoretical, barrett2021}), empirically demonstrating that both $\sigma_\phi$ and $\sigma$ reduce the Lipschitz constant of both the encoder and decoder networks and simultaneously improve the robustness of VAEs trained on CIFAR10.

\end{document}